%% file: arxiv-coxnam.tex
\documentclass{article}

\usepackage{microtype}
\usepackage{graphicx}
\usepackage{subcaption}
\usepackage{booktabs} 
\usepackage{multirow}
\usepackage[margin=0.8in]{geometry} 
\usepackage{tikz}
\usepackage{amsmath, amssymb}
\usepackage{pgfplots}
\usepackage{pgfplotstable}
\pgfplotsset{compat=1.18}
\usepackage{xcolor}
\usetikzlibrary{shapes.geometric, arrows, positioning, fit, backgrounds, calc, decorations.pathreplacing}
\usepackage{pdflscape}  
\usepackage{afterpage}  
\usepackage{natbib}
\usepackage{authblk}

\usepackage{hyperref}

\usepackage{mathtools}
\usepackage{amsthm}

\title{CRISP-NAM: Competing Risks Interpretable Survival Prediction with Neural Additive Models}

\author[1]{Dhanesh Ramachandram}
\author[1]{Ananya Raval}
\affil[1]{Vector Institute, Ontario, Canada}
\date{}

\begin{document}

\maketitle

\begin{abstract}
 Competing risks are crucial considerations in survival modelling, particularly in healthcare domains where patients may experience multiple distinct event types. We propose CRISP-NAM
 (Competing Risks Interpretable Survival Prediction with Neural Additive Models), an interpretable neural additive model for competing risks survival analysis which extends the neural additive architecture to model cause-specific hazards while preserving feature-level interpretability. Each feature contributes independently to risk estimation through dedicated neural networks, allowing for visualization of complex non-linear relationships between covariates and each competing risk.  CRISP-NAM demonstrates competitive performance on multiple datasets compared to existing approaches. 
  \end{abstract}

\section{Introduction}
\label{intro}
Survival analysis provides a robust framework for time-to-event prediction across multiple disciplines including medicine, finance, and manufacturing. Traditionally rooted in statistical methods, this analytical approach focuses on using available covariates to predict when specific events will occur. As an example, in healthcare settings, clinicians leverage patient data such as laboratory values, medical history, and comorbidities to forecast hospital readmission timing for specific medical conditions. 

Many real-world survival scenarios often involve competing risks where multiple mutually exclusive events can occur. While recent deep learning advances have improved predictive performance in competing risks settings, these models operate as ``black boxes''. This opacity poses significant challenges in high-stakes domains like healthcare, where understanding feature contribution to a model's predictions is crucial for clinical decision-making, regulatory compliance, and building trust with practitioners. 

Moreover, multiple jurisdictions have established interpretability requirements across general and sector-specific regulations. For example, Canada's Artificial Intelligence and Data Act (AIDA)~\citep{canada_aida_2024} emphasizes risk-based governance with interpretability assessments during development phases. Sector-specific regulations include Food and Drug Administration (FDA) guidance~\citep{fda_ai_software_medical_device} requiring clear documentation of AI/ML medical device decision-making and Article 13 of the EU AI Act~\citep{artificialintelligenceact_article13} which requires high risk AI systems to be designed with sufficient transparency and technical capabilities to explain their outputs, accompanied by clear instructions that enable users to properly interpret and appropriately use the system.

In this work, we bridge a critical gap between interpretability and competing risks modelling in deep survival analysis for healthcare applications by introducing CRISP-NAM (Competing Risks Interpretable Survival Prediction with Neural Additive Models). The main contributions of our work are as follows:

\begin{itemize}
\item We extend Neural Additive Models (NAMs) to competing risks settings, whereas existing NAM-based survival models only handle single-event outcomes.
\item We introduce separate projection functions for each risk-feature pair, allowing features to have differential effects across competing events while maintaining interpretability.
\item The proposed model jointly learns all cause-specific hazards in a single model, rather than treating competing events as censoring or requiring separate models.
\item Our model can be used to generate shape functions and feature importance rankings for each competing risk and this would allow practitioners to understand how different covariates influence each outcome.
\item We incorporate risk-frequency weightings to address class imbalances in competing events which is a common challenge that appears in real-world medical datasets.
\end{itemize}

The next section reviews the relevant background and related work that motivates our proposed approach.

\section{Background and Related Work}
\label{background}

\subsection{Cox Proportional Hazards Model}
Historically, the Cox Proportional Hazards (Cox PH) model~\citep{cox1972regression} has been a popular choice for survival analysis. It is a semi-parametric, linear model that relates covariates to the hazard function, which characterizes the instantaneous risk of an event occurring at time $t$, given survival up to that time. The Cox PH model assumes that covariates have a multiplicative effect on the hazard and that their effects are constant over time (\textit{proportional hazards assumption}). Mathematically, the hazard function under this model is expressed as:
\begin{equation}
h(t \mid X) = h_0(t) \exp(\beta^\top X)
\end{equation}
where $h_0(t)$ is the baseline hazard function, $X$ is the vector of covariates, and $\beta$ is the vector of regression coefficients.

Despite its widespread use and interpretability, the Cox PH model has several limitations. For example, nonlinear relationships must be manually specified (e.g., using splines), or alternatively introduced using nonlinear kernels~\cite{cai2011kernel} which can be challenging in high-dimensional settings. Additionally, this model assumes that the effect of each covariate on the hazard is constant over time. Violations of this assumption can lead to biased estimates. Finally, in scenarios with many features or complex interactions, prior feature engineering or dimensionality reduction may be necessary to avoid convergence issues or unstable estimates.

\subsection{Deep Learning for Survival Analysis}
To address the limitations of traditional statistical models, recent years have seen a shift towards machine learning-based survival models, including neural networks, which can capture nonlinear effects, interactions, and high-dimensional structure in the data. 

DeepSurv~\cite{katzman2018deepsurv} is one of the earliest deep learning models designed for survival analysis. It consists of a deep feedforward network with a single output node with a linear activation which estimates the log-risk function in the Cox PH model. \citet{kvamme2019time} introduced a joint time–covariate network $f(t, x)$, breaking the proportional hazards assumption by modelling the effect of $x$ as varying with time. This is conceptually closer to dynamic hazard models or time-dependent Cox PH models. 
With the aim of increasing trust and adoption, alignment with medical knowledge and supporting regulatory requirements, researchers have proposed several interpretable survival models in the literature. \citet{kovalev2020survlime} proposed SurvLIME, which incorporates the Local Interpretable Model-agnostic Explanation (LIME) framework~\cite{ribeiro2016model} to approximate the survival model in the local neighbourhood of a test instance in feature space.

Neural Additive Models (NAMs)~\cite{agarwal2021neural}, a neural-network extension of Generalized Additive Models (GAMs) have been used in machine learning based survival models, examples of which are SurvNAM~\cite{utkin2022survnam} and CoxNAM~\cite{xu2023coxnam}. While both SurvNAM and CoxNAM employ NAMs~\cite{agarwal2021neural}  to enhance interpretability in survival analysis, they differ fundamentally in purpose and integration. CoxNAM is a fully trainable survival model that embeds NAMs directly within the Cox proportional hazards framework, enabling inherently interpretable, end-to-end learning of nonlinear feature effects from survival data. In contrast, SurvNAM is a post-hoc explanation method that approximates the predictions of a pre-trained black-box survival model such as a Random Survival Forest~\citep{ishwaran2008random} by fitting a GAM-extended Cox PH model using NAMs as surrogate learners. 

Notably, none of the survival models discussed thus far are capable of modelling competing risks, which will be covered next.

\subsection{Competing Risks in Survival Analysis}
While conventional survival models primarily address single outcomes, many real-world scenarios  involve competing risks that fundamentally alter the probability distribution of the primary event. For instance, if a patient dies, the possibility of experiencing a subsequent heart‑related complication is removed, illustrating a typical competing‑events situation. Two methodological frameworks have emerged for analyzing competing risks:

The Cause-Specific Hazard approach~\citep{prentice1978analysis} models competing events separately, treating each outcome as a distinct hazard function and censoring subjects who experience competing events from the risk set, without requiring actual independence between event types. For each cause $k$, the cause-specific hazard $\lambda_k(t|\mathbf{x})$ represents the instantaneous rate of occurrence of event type $k$ at time $t$ for subjects who have not experienced any event prior to time $t$:
\begin{equation}
\small
\lambda_k(t|\mathbf{x}) = \lim_{\Delta t \to 0} \frac{P(t \leq T < t + \Delta t, E = k \mid \mathbf{x})}{\Delta t}
\end{equation}
where $\mathbf{x}$ represents the covariates,  $T$ represents the event time and $E \in \{1, 2, \ldots, K\}$ denotes the event type.

In contrast, the Fine-Gray sub-distribution model~\cite{fine1999proportional} directly accounts for competing events by maintaining subjects who experience competing risks within the risk set. Given the risk set for competing event $k$:
\begin{equation}
R_k^{sub}(t) = \{j: T_j \geq t \text{ or } (T_j < t \text{ and } E_j \neq k)\}
\end{equation}
This risk set includes subjects, $j$, who have either not experienced any event by time $t$ or have experienced a competing event (not event $k$) before time $t$.

The sub-distribution hazard is then defined as:
\begin{equation}
\small
\lambda_k^{sub}(t|\mathbf{x}) = \lim_{\Delta t \to 0} \frac{P(t \leq T < t + \Delta t, E = k | j \in R_k^{sub}(t), \mathbf{x})}{\Delta t}
\end{equation}

\subsection{Deep Survival Models for Competing Risks}
\begin{figure}
    \centering
    \includegraphics[width=0.75\linewidth]{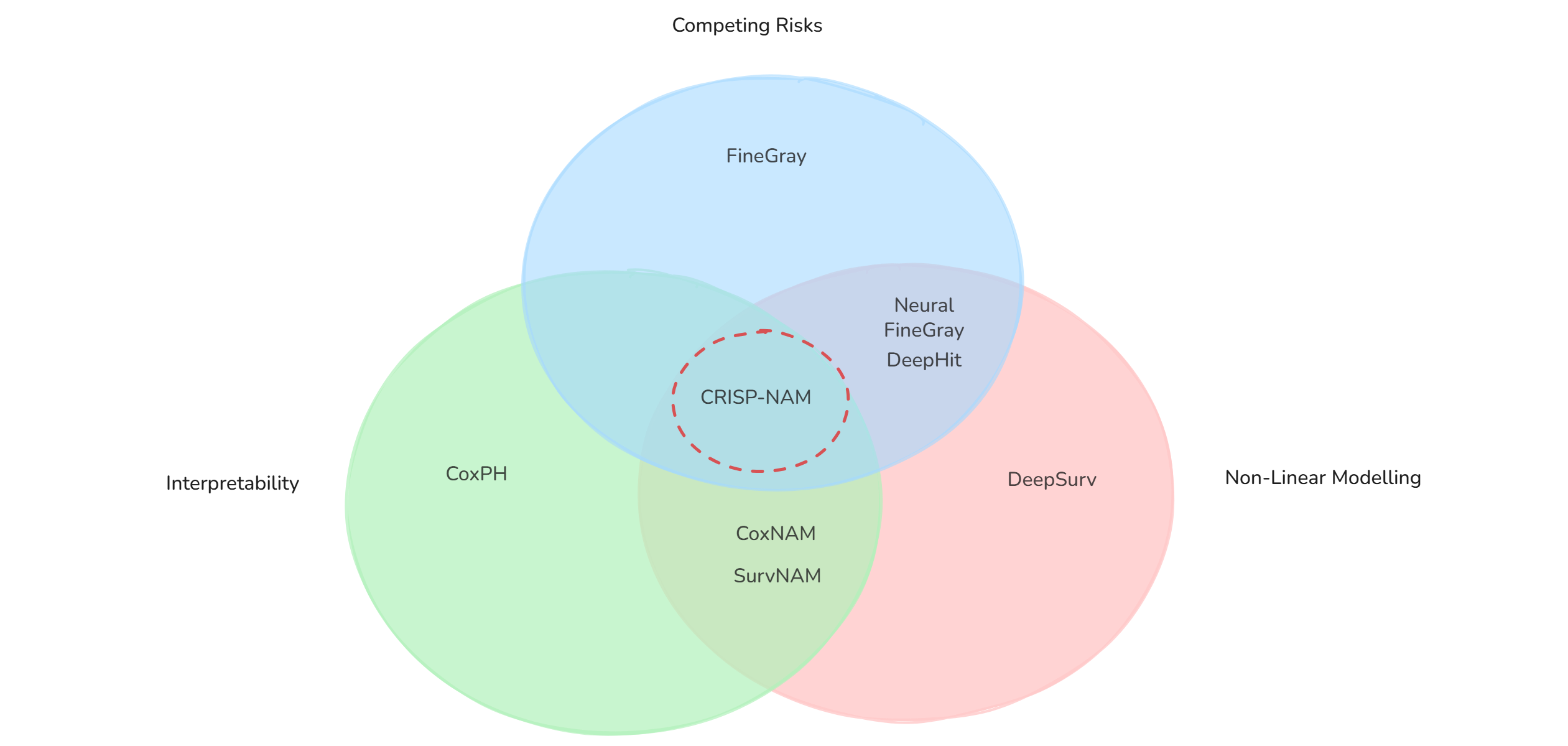}
    \caption{Venn diagram situating CRISP-NAM against existing deep survival methods}
    \label{fig:venn}
\end{figure}

Deep Survival Models leverage deep learning techniques to address competing risks in survival analysis, offering the ability to model complex non-linear patterns in risk prediction. DeepHit~\cite{lee2018deephit} is a joint model for survival analysis with competing risks. It uses a shared representation network followed by cause-specific sub-networks to model the joint distribution of the event time and event type. The model is trained using a combination of the negative log-likelihood and a ranking loss to encourage concordance between predicted risks and observed outcomes. Neural Fine Gray~\cite{jeanselme2023neural} extends the Fine-Gray sub-distribution model using neural networks to capture non-linear relationships between covariates and sub-distribution hazards. It allows for flexible modelling of competing risks while maintaining the ability to directly estimate cumulative incidence functions.
Despite these advances, a key limitation of existing deep survival approaches for competing risks is their lack of interpretability, especially at the feature level, making it difficult to understand how individual features contribute to risk predictions for different competing events. In Fig.~\ref{fig:venn}, we depict the current gap in the literature of deep survival models. Specifically, we are interested in these criteria: Interpretability, Non-Linear Modelling and Competing Risks Capability.
Models such as SurvNAM and CoxNAM are interpretable, however, they have not been reported to be used in competing risks settings. In contrast, the Neural Fine-Gray and DeepHit architectures can model competing risks, but are ``black-box'' models and can only be explained using post-hoc explainability methods such as 
SHAP and Partial Dependence Plots. Post hoc methods are known to be imprecise and can generate misleading explanations~\cite{huang2024shapley, dangers-post-hoc}. The original formulations of Fine-Gray and Cox PH models are  incapable of modelling non-linear relationships. Our proposed CRISP-NAM model addresses these gaps in survival models fulfilling all 3 criteria while providing competitive performance. 

To this end, an extension of the Neural Additive Model for competing risks settings is proposed in this paper, resulting in an inherently interpretable survival model. Our approach retains the interpretability and feature-wise transparency of NAMs and allowing for flexible, non-linear modelling of cause-specific or sub-distribution hazards in competing risks scenarios.

\section{Model Architecture}
CRISP-NAM extends Neural Additive Models to the competing risks survival analysis setting while preserving feature-level interpretability. 
The architecture consists of three primary components:
\begin{figure*}[ht]
    \centering
    \resizebox{\textwidth}{!}{
        \input{arch-diag}
    }
    \caption{CRISP-NAM architecture showing the flow from input features through neural networks for two competing risk predictions.}
    \label{fig:crisp-nam-arch}
\end{figure*}

\subsection{Neural Additive Model (FeatureNet)}
In line with the Neural Additive Model framework, each input feature $x_i$ is processed by its own dedicated neural network $f_i(\cdot)$, referred to here as a \emph{FeatureNet}. These feature-specific sub-networks are designed to learn the non-linear contribution of each individual feature to the overall risk score, while preserving interpretability by isolating feature effects. 
\begin{equation}
\mathbf{h}_i = f_i(x_i) \in \mathbb{R}^d
\end{equation}
where $d$ is the dimension of the hidden representation. Each FeatureNet is a fully-connected feedforward neural network with $L$ layers, taking the scalar input $x_i$ and producing a hidden representation $\mathbf{h}_i \in \mathbb{R}^d$. The activations are computed recursively using the hyperbolic tangent function:

\begin{equation}
\mathbf{z}^{(l)}_i = \tanh(W^{(l)}_i \mathbf{z}^{(l-1)}_i + b^{(l)}_i), \quad \text{for } l = 1, \ldots, L,
\end{equation}

where $\mathbf{z}^{(0)}_i = x_i$, and $\mathbf{h}_i = \mathbf{z}^{(L)}_i$ denotes the output of the final layer. Each layer $l$ has weights $W^{(l)}i \in \mathbb{R}^{d_l \times d{l-1}}$ and biases $b^{(l)}_i \in \mathbb{R}^{d_l}$, with $d_0 = 1$ and $d_L = d$.

In this implementation, Dropout is used with rate $p_{\text{dropout}}$ after each hidden layer, Feature Dropout with rate $p_{\text{feature}}$ during training to increase robustness and an optional batch normalization layer after each linear transformation to stabilize learning especially for deeper FeatureNets.

\subsection{Risk-Specific Projections}

For each feature $i$ and competing risk $k$, a separate linear projection transforms the feature representation to its contribution to the log-hazard ratio. To address scale ambiguities across different competing risks and ensure fair comparison of feature contributions, we constrain the projection vectors to have unit L2 norm.

The risk-specific projection is defined as:
\begin{equation}
g_{i,k}(\mathbf{h}_i) = \tilde{\mathbf{w}}_{i,k}^T \mathbf{h}_i
\end{equation}
where $\tilde{\mathbf{w}}_{i,k}$ is the L2-normalized projection vector:
\begin{equation}
\tilde{\mathbf{w}}_{i,k} = \frac{\mathbf{w}_{i,k}}{\|\mathbf{w}_{i,k}\|_2 + \epsilon}
\end{equation}
with $\mathbf{w}_{i,k} \in \mathbb{R}^d$ being the learnable weight vector for projection $i \in \{1, 2, \ldots, p\}$ and risk $k \in \{1, 2, \ldots, K\}$, and $\mathbf{h}_i \in \mathbb{R}^d$ being the feature representation.

This normalization constraint ensures that $\|\tilde{\mathbf{w}}_{i,k}\|_2 = 1$ for all feature-risk pairs, which constraints all projection vectors to operate on the same scale and enabling direct comparison of feature importance across different competing risks.  

\subsection{Additive Risk Aggregation}

The cause-specific log-hazard ratio for risk $k$ given input features $\mathbf{x} = [x_1, x_2, \ldots, x_p]$ is computed as the sum of individual feature contributions:
\begin{equation}
\eta_k(\mathbf{x}) = \sum_{i=1}^{p} g_{i,k}(f_i(x_i))
\end{equation}
This preserves the additive nature of the model while allowing for complex non-linear feature effects.

\subsection{Cause-Specific Hazards Approach}

 The cause-specific hazards framework~\cite{prentice1978analysis} is adopted for the CRISP-NAM model. For a subject with covariates $\mathbf{x}$, we parameterize each cause-specific hazard using a Cox-type model:
\begin{equation}
\lambda_k(t|\mathbf{x}) = \lambda_{0k}(t)\exp(\eta_k(\mathbf{x}))
\end{equation}
where $\lambda_{0k}(t)$ is the baseline hazard for the $k$-th event and $\eta_k(\mathbf{x})$ is the risk score function for event type $k$. Unlike traditional Cox models with linear risk functions, our approach uses FeatureNets within a neural additive model (NAM), enabling it to model complex non-linear effects of individual features.

\subsection{Partial Likelihood Loss Function}
To train the model, the standard Cox partial likelihood approach, adapted for competing risks~\cite{prentice1978analysis}, is implemented. For a dataset with $N$ subjects, the negative log partial likelihood for event type $k \in \{1,\dots,K\}$ is
\begin{equation}
\small
\mathcal{L}_k = -\sum_{\substack{n=1 \\ E_n = k}}^{N} \Bigg[ \eta_k(\mathbf{x}_n) - 
\log\left(\sum_{\substack{j=1 \\ T_j \geq T_n}}^{N} \exp(\eta_k(\mathbf{x}_j))\right) \Bigg]
\end{equation}
where $E_n$ is the event type for subject $n$ and $T_n$ is their event or censoring time. The risk set at time $T_n$ consists of all subjects $j$ who have not yet experienced any event ($T_j \geq T_n$), and $\mathbf{x}_j$ denotes the feature vector for each subject $j$ in this risk set.
The overall loss is the sum of the negative log partial likelihoods across all event types:
\begin{equation}
\mathcal{L} = \sum_{k=1}^{K} \mathcal{L}_k + \gamma \|\Theta\|^2_2
\end{equation}
where $\gamma$ is the $L_2$ regularization parameter and $\Theta$ represents all model parameters.
Since many real-world problems involving competing risks suffer from class imbalance, 
we adopt a risk-frequency-weighted version of the partial likelihood. Specifically, we define:
\begin{align}
\small
\mathcal{L}_{k,\omega} = 
    - \omega_k \sum_{\substack{n=1 \\ E_n = k}}^{N} \bigg[
        \eta_k(\mathbf{x}_n)
        - \log \Big(
            \sum_{\substack{j=1 \\ T_j \geq T_n}}^{N} \exp\big( \eta_k(\mathbf{x}_j) \big)
        \Big)
    \bigg]
\end{align}
where $\omega_k$ is a weight inversely proportional to the frequency of event type $k$. 
The total loss is given by:
\begin{equation}
\mathcal{L}_{\text{weighted}} = \sum_{k=1}^{K} \mathcal{L}_{k,\omega}
\end{equation}

\subsection{Baseline Hazard Estimation}

In the cause-specific proportional hazards formulation, the hazard for event type $k \in \{1,\dots,K\}$ at time $t \geq 0$ for a subject with covariates $\mathbf{x} \in \mathbb{R}^p$ is expressed as
\begin{equation}
\lambda_k(t \mid \mathbf{x}) = \lambda_{0k}(t) \cdot \exp\!\big( \eta_k(\mathbf{x}) \big),
\end{equation}
where 
\begin{itemize}
\item $\lambda_{0k}(t)$ is the baseline cause-specific hazard function for event type $k$, independent of covariates,
\item $\eta_k(\mathbf{x})$ is the covariate-dependent log-risk function produced by the model.
\end{itemize}

We do not directly parameterize $\lambda_{0k}(t)$. Instead, we estimate the corresponding baseline cumulative hazard function $\lambda_{0k}(t) = \int_0^t \lambda_{0k}(u)\,du$ after training using the Breslow estimator~\cite{breslow1975analysis}:
\begin{equation}
\hat{\lambda}_{0k}(t_m)
= \sum_{\substack{n: T_n \leq t \\ E_n = k}} 
\frac{1}{\sum_{j: T_j \geq T_i} \exp\!\left(\eta_k(\mathbf{x}_j)\right)},
\end{equation}
where $T_n$ denotes the observed time for subject $n$, $E_n \in \{0,1,\dots,K\}$ is the event indicator ($E_n=0$ if censored), and the denominator represents the sum of relative risks for all subjects still at risk at time $T_n$. 

This estimated baseline cumulative hazard $\hat{\lambda}_{0k}(t_m)$ can then be used to compute the cumulative incidence function (CIF) for each event type $k$ at test time.

Estimating baseline hazards is necessary for several reasons. First, while the neural additive component of CRISP-NAM efficiently learns relative risks between subjects (hazard ratios), baseline hazard estimation enables translation of these relative measures into absolute risk predictions. This is required for clinical decision-making, where probabilities of events are needed. Second, in competing risks settings, accurate baseline hazard estimation is required for proper calculation of CIFs, as shown in Eqs.~\eqref{eq:CIF} and~\eqref{Eq:surv_func}. Third, the baseline hazard captures the underlying temporal pattern of risk independent of covariates, allowing CRISP-NAM to generate time-dependent predictions at clinically relevant horizons (e.g., 1-year, 5-year risks). Finally, proper evaluation metrics such as Brier scores and time-dependent AUCs at specific time points depend on accurate absolute risk estimation.

\subsection{Prediction of Absolute Risks}

To predict the cumulative incidence function (CIF)~\citep{gray1988class} for each competing event, we use the relationship between cause-specific hazards and the CIF. For a subject with covariates $\mathbf{x}$, the CIF for event type $k$ at time $t \geq 0$ is defined as
\begin{equation}
\label{eq:CIF}
F_k(t|\mathbf{x}) = \int_0^t S(u|\mathbf{x}) \,\lambda_k(u|\mathbf{x}) \, du,
\end{equation}
where 
\begin{itemize}
\item $S(t|\mathbf{x})$ is the overall survival function, i.e., the probability of not experiencing any event up to time $t$, 
\item $u$ is the integration variable representing time between $0$ and $t$, 
\item $\lambda_k(u|\mathbf{x}) = \lambda_{0k}(u)\exp(\eta_k(\mathbf{x}))$ is the cause-specific hazard for cause $k$.
\end{itemize}

The survival function is given by
\begin{equation}
\label{Eq:surv_func}
S(t|\mathbf{x}) = \exp\!\left(-\sum_{k=1}^{K} \int_0^t \lambda_k(u|\mathbf{x}) \, du\right).
\end{equation}

In practice, a discrete approximation is used to compute these integrals. Let $\{t_1, t_2, \dots, t_M\}$ denote a set of ordered discrete time points with $t_m \in [0,t]$. Then the CIF can be approximated as
\begin{equation}
\label{eq:discrete_cif}
\hat{F}_k(t|\mathbf{x}) \approx \sum_{t_m \leq t} \hat{S}(t_{m-1}|\mathbf{x}) \cdot \hat{\lambda}_k(t_m|\mathbf{x}),
\end{equation}
where 
\begin{align*}
\hat{S}(t_{m-1}|\mathbf{x}) &= \exp\!\left( -\sum_{k'=1}^K \sum_{t_\ell < t_m} \hat{\lambda}_{k'}(t_\ell|\mathbf{x}) \right), \\
\hat{\lambda}_k(t_m|\mathbf{x}) &= \hat{\lambda}_{0k}(t_m) \exp\!\big(\eta_k(\mathbf{x})\big),
\end{align*}
and $\hat{\lambda}_{0k}(t_m)$ denotes the estimated baseline cause-specific hazard at time $t_m$ for event type $k$.


\subsection{Interpretability Mechanisms}

Given that CRISP-NAM is based on NAMs, as with all variants of Generalized Additive Models,  CRISP-NAM can generate \emph{shape functions plots} to visualize the (non-linear) contribution of each feature to the prediction. Specifically, for each feature $i$ and risk $k$, we can extract a shape function that describes how the feature affects the log-hazard ratio.

\begin{equation}
s_{i,k}(x_i) = g_{i,k}(f_i(x_i))
\end{equation}

The importance of feature $i$ for risk $k$ is quantified by the mean absolute value of its contribution across the dataset.
\begin{equation}
\mathcal{I}_{i,k} = \frac{1}{N} \sum_{j=1}^{N} \left| s_{i,k}(x_{ij}) \right|
\end{equation}

This enables ranking features by their impact on each competing risk, providing valuable insights into risk-specific predictor importance.

Notably, in this current implementation, CRISP-NAM does not capture features interactions. This is by design to prioritize interpretability through independent feature level shape functions, ensuring that feature contributions can be visualized and understood in isolation. Adding separate FeatureNets to model feature interactions adds to the model complexity and affects its interpretability as visualization beyond pairwise features interactions is challenging. With $p$ features, there are $\mathcal{O}(p^2)$ possible pairwise interactions, and deciding which interactions to model would also require domain knowledge.

\section{Experiments}

In this section, the datasets used and the experimental procedure to evaluate the CRISP-NAM model are described.

\subsection{Datasets}
In order to evaluate the proposed interpretable model for competing risks survival prediction, we used the following three real-world medical datasets and a synthetic dataset. Table~\ref{tab:dataset_summary} provides a summary and breakdown of the primary and competing risks for each the datasets used in this work.

\paragraph{Primary Biliary Cholangitis (PBC).} The PBC dataset originates from a randomized controlled trial conducted at the Mayo Clinic between 1974 and 1984, involving 312 patients diagnosed with primary biliary cholangitis. The study aimed to evaluate the efficacy of D-penicillamine in treating the disease. Each patient record includes 25 covariates encompassing demographic, clinical, and laboratory measurements. The primary endpoint was mortality while on the transplant waiting list, with liver transplantation considered a competing risk \citep{Dickson1989}.

\paragraph{Framingham Heart Study.} Initiated in 1948, the Framingham Heart Study is a longitudinal cohort study designed to investigate cardiovascular disease (CVD) risk factors. For this analysis, data from 4,434 male participants were utilized, each with 18 baseline covariates collected over a 20-year follow-up period. The study focuses on modelling the risk of developing CVD, treating mortality from non-CVD causes as a competing event~\cite{kannel1979diabetes}.

\paragraph{SUPPORT2 Dataset.} The SUPPORT2 dataset originates from the Study to Understand Prognoses and Preferences for Outcomes and Risks of Treatments (SUPPORT2), a comprehensive investigation conducted across five U.S. medical centers between 1989 and 1994. This dataset encompasses records of 9,105 critically ill hospitalized adults, each characterized by 42 variables detailing demographic information, physiological measurements, and disease severity indicators. The study was executed in two phases: Phase I (1989–1991) was a prospective observational study aimed at assessing the care and decision-making processes for seriously ill patients and Phase II (1992–1994) implemented an intervention to enhance end-of-life care. The primary objective was to develop and validate prognostic models estimating 2- and 6-month survival probabilities, thereby facilitating improved clinical decision-making and patient-physician communication regarding treatment preferences and outcomes \citep{support1995controlled}.

\paragraph{Synthetic Dataset.}
We use the synthetic dataset introduced by Lee et al.~\cite{lee2018deephit}, which models two competing risks with distinct but overlapping covariate effects. Each patient \( i \) is assigned a 12-dimensional feature vector \( \mathbf{x}^{(i)} \sim \mathcal{N}(0, I_{12}) \), partitioned into three 4-dimensional subgroups: \( \mathbf{x}_1^{(i)}, \mathbf{x}_2^{(i)}, \mathbf{x}_3^{(i)} \). The event times are sampled from exponential distributions as:
\begin{align}
T_1^{(i)} &\sim \exp\left( \gamma_T \|\mathbf{x}_3^{(i)}\|^2 + \gamma_T \mathbf{1}^\top \mathbf{x}_1^{(i)} \right), \\
T_2^{(i)} &\sim \exp\left( \gamma_T \|\mathbf{x}_3^{(i)}\|^2 + \gamma_T \mathbf{1}^\top \mathbf{x}_2^{(i)} \right),
\end{align}
where \( \gamma_T = 10 \). Covariates \( \mathbf{x}_1 \) and \( \mathbf{x}_2 \) influence only their respective event times, while \( \mathbf{x}_3 \) affects both.

The dataset consists of 30{,}000 rows of unique patient data with 50\% random right-censoring by drawing a censoring time \( t_c^{(i)} \sim \mathcal{U}[0, \min\{T_1^{(i)}, T_2^{(i)}\}] \). The final observed data for each patient is \( (\mathbf{x}^{(i)}, t^{(i)}, k^{(i)}) \), where \( t^{(i)} \) is the observed time and \( k^{(i)} \) is the event indicator (\( \emptyset \) if censored).

\begin{table}[htbp]
\centering
\caption{Dataset characteristics and competing risk statistics}
\label{tab:dataset_summary}
\begin{tabular}{@{}lrrllr@{}}
\toprule
Dataset & Observations & Features & Primary & Competing Risk & Censored \\
\midrule
PBC & 312 & 25 & Death (44.87\%) & Transplant (9.29\%) & 45.83\% \\
Framingham & 4,434 & 18 & CVD (26.09\%) & Death (17.75\%) & 56.16\% \\
SUPPORT2 & 9,105 & 42 & Cancer\_Death (18.2\%) & Death\_Other (49.9\%) & 31.9\% \\
Synthetic & 30,000 & 12 & * (25.33\%) & * (24.67\%) & 50.00\% \\

\bottomrule
\end{tabular}
\end{table}

\subsection{Experimental Setup} 

The CRISP-NAM model is implemented using PyTorch and the code is available from \url{https://github.com/VectorInstitute/crisp-nam}. We employed nested cross-validation to prevent data leakage during hyperparameter optimization and model evaluation. The approach consists of an outer 5-fold stratified cross-validation for performance evaluation and an inner 5-fold cross-validation for hyperparameter tuning within each outer fold. For each outer fold, Optuna~\citep{optuna} is employed on the training partition to systematically search for optimal model configurations using the inner 5-fold cross-validation, tuning learning rate, $L_2$ regularization strength, dropout rates, network architecture (1-3 hidden layers with 8-128 units), and batch normalization settings using validation loss as the objective. The best configuration identified for each outer fold is then trained on the complete training partition and evaluated on the corresponding held-out test partition.  Continuous features were normalized using standard scaling ($\mu=0, \sigma=1$) and categorical features were one-hot encoded. Missing categorical values were imputed using mode imputation. For continuous variables, mean imputation was used across all datasets. Training employed the \textit{AdamW}~\citep{adamw2017} optimizer to minimize the negative log-likelihood loss with a batch size of 256 and early stopping (patience=10) to prevent over-fitting.

Model performance was assessed using complementary metrics~\citep{park2021review} for discrimination and calibration. For discrimination ability, the Time-Dependent Area-Under-the-Curve (TD-AUC) was used to quantify how well the model ranks subjects by risk, with values ranging from 0.5 (no better than chance) to 1.0 (perfect discrimination). Additionally, the Time-Dependent Concordance Index (TD-CI), which considers that the model's performance can change over time was computed. This is crucial because the risk of an event may evolve as time progresses, and a model's predictive ability might not be constant. TD-CI ranges from 0.5 to 1.0, with higher values indicating better discriminative ability. The third metric was the Brier score (BS), which measures the accuracy of probabilistic predictions and penalizes both discrimination and calibration errors with lower values of this score indicating better performance. All metrics were evaluated at multiple clinically relevant time horizons corresponding to the 25th, 50th, and 75th percentiles of observed event times for each competing risk.

\section{Results and Discussion}

\begin{table*}[ht]
\caption{Comparative 5-fold performance metrics for various models across multiple competing risks datasets.}
\label{Tab:results}
\vskip 0.15in
\begin{center}
\begin{scriptsize}
\resizebox{\textwidth}{!}{
\begin{tabular}{llc@{\hspace{0.5em}}ccc@{\hspace{0.7em}}ccc@{\hspace{0.7em}}ccc}
\toprule
\textbf{Dataset} & \textbf{Model} & \textbf{Risk} & \multicolumn{3}{c}{\textbf{TD-AUC}} & \multicolumn{3}{c}{\textbf{TD-CI}} & \multicolumn{3}{c}{\textbf{Brier Score}} \\
\cmidrule(lr){4-6} \cmidrule(lr){7-9} \cmidrule(lr){10-12}
& & & $q_{.25}$ & $q_{.50}$ & $q_{.75}$ & $q_{.25}$ & $q_{.50}$ & $q_{.75}$ & $q_{.25}$ & $q_{.50}$ & $q_{.75}$ \\
\midrule
\multirow{6}{*}{FHS} 
& \multirow{2}{*}{CRISP-NAM} & 1 & 0.843±.021 & \textbf{0.832±.028} & \textbf{0.811±.040} & 0.708±.022 & 0.700±.021 & 0.691±.023 & 0.249±.295 & 0.320±.273 & 0.353±.247 \\
& & 2 & 0.793±.050 & \textbf{0.779±.027} & 0.771±.028 & 0.714±.029 & 0.713±.017 & 0.707±.014 & 0.041±.004 & 0.079±.006 & 0.127±.010 \\
\cmidrule(lr){2-12}
& \multirow{2}{*}{NFG} & 1 & 0.678±.160 & 0.673±.149 & 0.666±.133 & 0.632±.150 & 0.629±.141 & 0.628±.134 & 0.058±.005 & 0.102±.005 & \textbf{0.134±.007} \\
& & 2 & 0.617±.126 & 0.629±.117 & 0.620±.136 & 0.612±.156 & 0.611±.148 & 0.610±.152 & 0.041±.003 & 0.077±.006 & 0.113±.009 \\
\cmidrule(lr){2-12}
& \multirow{2}{*}{DeepHit}  & 1 & \textbf{0.854±.019} & 0.831±.012 & 0.807±.009 & \textbf{0.738±.030} & \textbf{0.729±.018} & \textbf{0.724±.017} & \textbf{0.056±.006} & \textbf{0.101±.004} & 0.134±.004 \\
& & 2 & \textbf{0.796±.053 }& 0.779±.028 & \textbf{0.776±.031} & \textbf{0.737±.030} & \textbf{0.728±.018} & \textbf{0.724±.017} & \textbf{0.038±.003} & \textbf{0.072±.004} & \textbf{0.109±.005} \\
\midrule
\multirow{6}{*}{SUP} 
& \multirow{2}{*}{CRISP-NAM} & 1 & \textbf{0.855±.065} & \textbf{0.802±.092} & \textbf{0.798±.100} & 0.665±.036 & \textbf{0.624±.029} & \textbf{0.617±.028} & 0.277±.268 & 0.380±.242 & 0.393±.227 \\
& & 2 & 0.872±.188 & 0.838±.179 & 0.813±.175 & \textbf{0.804±.110} & \textbf{0.713±.097} & \textbf{0.680±.087} & 0.287±.266 & 0.308±.188 & 0.308±.129 \\
\cmidrule(lr){2-12}
& \multirow{2}{*}{NFG}       & 1 & 0.790±.036 & 0.777±.023 & 0.797±.020 & 0.620±.038 & 0.513±.025 & 0.438±.025 & 0.044±.003 & 0.100±.005 & 0.115±.005 \\
& & 2 & 0.902±.004 & 0.840±.003 & 0.812±.005 & 0.847±.008 & 0.741±.007 & 0.701±.010 & 0.092±.004 & 0.151±.004 & 0.176±.002 \\
\cmidrule(lr){2-12}
& \multirow{2}{*}{DeepHit}  & 1 & 0.779±.205 & 0.640±.222 & 0.665±.188 & \textbf{0.746±.013} & 0.540±.010 & 0.497±.011 & \textbf{0.039±.003} & \textbf{0.089±.002} & \textbf{0.101±.003} \\
& & 2 & \textbf{0.955±.025} & \textbf{0.893±.069} & \textbf{0.860±.083} & 0.863±.008 & 0.745±.005 & 0.697±.006 & \textbf{0.087±.005 }& \textbf{0.144±.006} & \textbf{0.171±.003} \\
\midrule
\multirow{6}{*}{PBC} 
& \multirow{2}{*}{CRISP-NAM} & 1 & 0.988±.012 & 0.958±.028 & 0.942±.034 & 0.804±.019 & 0.785±.011 & 0.773±.006 & 0.131±.023 & 0.184±.042 & 0.242±.074 \\
& & 2 & 0.952±.043 & 0.966±.023 & 0.972±.017 & 0.647±.009 & 0.635±.015 & 0.613±.016 & 0.469±.223 & 0.514±.217 & 0.537±.217 \\
\cmidrule(lr){2-12}
& \multirow{2}{*}{NFG}       & 1 & 0.853±.023 & 0.835±.011 & 0.824±.027 & 0.782±.017 & 0.765±.013 & 0.756±.016 & 0.143±.020 & 0.152±.008 & 0.170±.013 \\
& & 2 & 0.491±.055 & 0.491±.055 & 0.537±.056 & 0.227±.022 & 0.238±.011 & 0.245±.016 & 0.092±.054 & 0.135±.073 & 0.164±.082 \\
\cmidrule(lr){2-12}
& \multirow{2}{*}{DeepHit}  & 1 & \textbf{0.994±.008} & \textbf{0.965±.028} & \textbf{0.959±.034} & \textbf{0.822±.016} & \textbf{0.796±.019} & \textbf{0.776±.013} & \textbf{0.100±.009} & \textbf{0.130±.004} &\textbf{ 0.152±.011} \\
& & 2 & \textbf{0.978±.036} & \textbf{0.983±.027} & \textbf{0.987±.020} & \textbf{0.728±.024} & \textbf{0.705±.019 }& \textbf{0.690±.016} & \textbf{0.037±.006} & \textbf{0.056±.003} & \textbf{0.065±.004} \\
\midrule

\multirow{6}{*}{SYN} 
& \multirow{2}{*}{CRISP-NAM} & 1 & 0.655±.019 & 0.660±.041 & 0.670±.081 & 0.551±.011 & 0.559±.008 & 0.555±.004 & 0.245±.336 & 0.301±.287 & 0.347±.219 \\
& & 2 & 0.676±.026 & 0.653±.022 & 0.643±.021 & 0.569±.014 & 0.564±.014 & 0.560±.014 & 0.226±.246 & 0.356±.245 & 0.433±.172 \\
\cmidrule(lr){2-12}
& \multirow{2}{*}{NFG}       & 1 & 0.802±.011 & 0.771±.016 & 0.717±.016 & 0.741±.007 & 0.715±.008 & 0.670±.006 & \textbf{0.051±.001} & \textbf{0.096±.002} & 0.161±.003 \\
& & 2 & 0.815±.013 & 0.771±.016 & 0.717±.019 & 0.739±.024 & 0.706±.026 & 0.662±.023 & 0.049±.002 & \textbf{0.095±.003} & 0.160±.003 \\
\cmidrule(lr){2-12}
& \multirow{2}{*}{DeepHit}    & 1 & \textbf{0.847±.006} & \textbf{0.833±.010} & \textbf{0.815±.006} & \textbf{0.757±.006} & \textbf{0.734±.007} & \textbf{0.696±.004} & 0.052±.001 & 0.097±.002 & \textbf{0.160±.002} \\
& & 2 & \textbf{0.855±.009} & \textbf{0.829±.010} & \textbf{0.810±.012} & \textbf{0.762±.009} & \textbf{0.737±.008} & \textbf{0.695±.007} & \textbf{0.049±.001} & \textbf{0.095±.003} & \textbf{0.159±.001} \\
\bottomrule
\end{tabular}
}
\end{scriptsize}
\end{center}
\vskip 0.1in
\footnotesize
\textbf{Notes:} Dataset = (FHS: Framingham Heart Study, SUP: SUPPORT2, PBC: Primary Biliary Cirrhosis, SYN: Synthetic); 
Model= (CRISP-NAM: CRISP-NAM, NFG: Neural Fine Gray, DEEPHIT: DeepHit); 
Risk =  (1: Primary, 2: Competing)
\end{table*}


Table~\ref{Tab:results} displays the 5-fold cross-validated performance of CRISP-NAM against two state-of-the-art neural baselines: DeepHit and Neural Fine Gray (NFG). While DeepHit generally achieves the highest performance, CRISP-NAM demonstrates competitive discrimination with the added benefit of interpretability through its additive structure.

CRISP-NAM achieves discrimination metrics within 2-5\% of DeepHit on clinical datasets, demonstrating its competitiveness. On FHS, the TD-AUC gap at $q_{0.25}$ is merely 0.011 (0.843 vs 0.854) and CRISP-NAM performance is at parity with DeepHit at $q_{0.50}$ for both risks. Similarly, on SUPPORT2 Risk 1, CRISP-NAM exceeds DeepHit's TD-AUC by substantial margins at $q_{0.25}$ (0.855 vs 0.779) and $q_{0.50}$ (0.802 vs 0.640). For PBC, while DeepHit achieves marginally higher TD-AUC, both models reach near-ceiling performance, making the practical difference negligible. The synthetic dataset represents the primary challenge, where DeepHit's flexible architecture captures complex non-linear patterns that CRISP-NAM's additive structure cannot fully represent. Our experimental results demonstrate a systematic 15\% performance deficit of CRISP-NAM relative to DeepHit across all metrics on the synthetic dataset. The synthetic dataset generator incorporates quadratic univariate effects through ${||x_3||}^2$, which creates strong non-linearities in the hazard space that may be challenging for CRISP-NAM's additive architecture. The $x_3$ components also affect both competing risks identically,  creating  dependencies that are difficult for strictly additive models to capture by the CRISP-NAM model.

Notably, CRISP-NAM's calibration performance (Brier score) is also lower than NFG and DeepHit. This could be attributed to the loss function we implemented that optimizes for discrimination (ranking) rather than the calibration (probability accuracy). In addition, the event weighting could exacerbate the model's lagging calibration. We will investigate improvements to the loss function to handle calibration more effectively in our future work.

\subsection{Interpretability Analysis}

Here, the shape function plots for the top 10 features per risk as learned by the CRISP-NAM model across 3 datasets: SUPPORT2, Framingham and PBC are presented. Each curve $s_{i,k}(x_i)$ represents the marginal contribution of feature $x_i$ to the log cause-specific hazard for risk $k$. Rug plots beneath each curve illustrate the empirical distribution of feature values, highlighting regions, particularly in the tails, where data are sparse.

\textbf{Interpretation Note.} It should be noted that these shape functions are \emph{associational}, not causal, and may obscure interactions between features. They are estimated under smoothness constraints and can extrapolate in regions with low data density, leading to amplified or flattened effects.  Apparent patterns should be interpreted cautiously, corroborated on external cohorts, and discussed with domain experts before drawing scientific or clinical conclusions. While our primary focus centres on analyzing trends revealed through the shape function plots, we provide limited discussion of the associations between covariates and predicted risks. These discussions serve primarily to demonstrate how our shape plot findings align with or contrast against established medical literature.

\subsubsection{Framingham Heart Study Dataset}

\begin{figure}[htbp]
    \centering
    \begin{subfigure}[b]{\textwidth}
        \includegraphics[width=\linewidth, height=8cm, keepaspectratio]{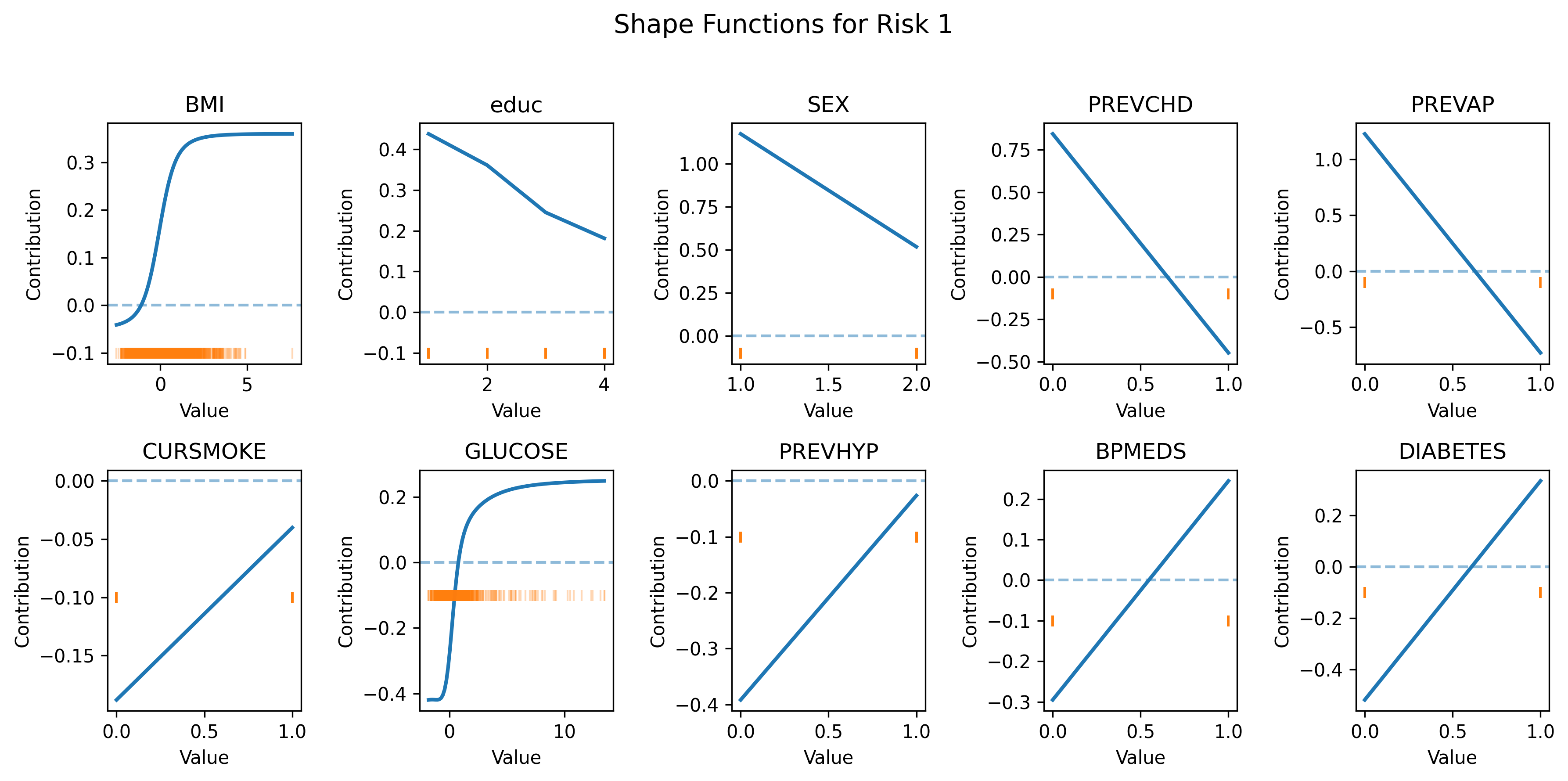}
    \end{subfigure}
    \vspace{1em}
    \begin{subfigure}[b]{\textwidth}
        \includegraphics[width=\linewidth, height=8cm, keepaspectratio]{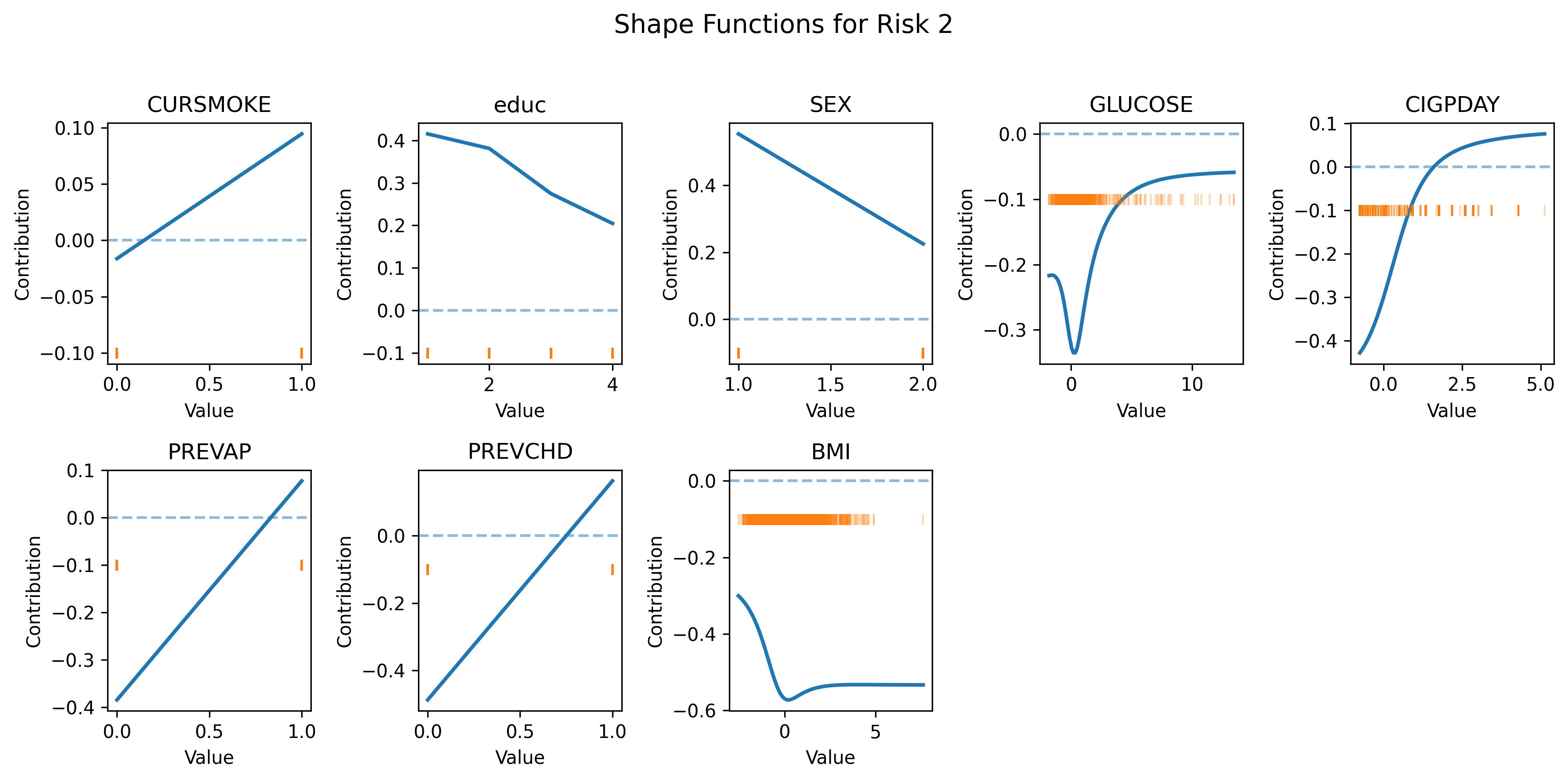}
    \end{subfigure}

    \caption{Shape functions computed with CRISP-NAM model for the top-10 important features: Framingham dataset}
    \label{Fig:Framingham_shape}
\end{figure}

Figure~\ref{Fig:Framingham_shape} illustrates shape functions from the CRISP-NAM model trained on the Framingham dataset, with separate plots for two competing risks: cardiovascular disease (\textit{Risk 1}) and non–cardiovascular death (\textit{Risk 2}). All functions are displayed on the log-hazard scale, where a unit increase of $+0.69$ corresponds to a doubling of the cause-specific hazard.

For \textit{Risk 1}, \texttt{GLUCOSE} and \texttt{BMI} shows steep increase showing that higher blood glucose levels and increasing BMI are associated to higher risk of cardiovascular disease for this dataset. Shape plots reveal that the patients with diabetes, patients who are on medication for hypertension and who are current smokers (\texttt{DIABETES}, \texttt{BPMEDS} and \texttt{CURSMOKE}) have higher risk of cardiovascular disease which correlates well with known risk factors. Shape plot also reveals that women have lower risk of cardiovascular disease, compared to men and having more education is correlated with lower risk of cardiovascular disease. The binary history flags \texttt{PREVCHD} and \texttt{PREVAP} show negative contributions to the log-hazard. Three data characteristics, rather than a real reversal of risk, explain this result:

\begin{enumerate}
    \item \textbf{Sparsity.} For binary variables with low prevalence rates (less than 10\%) such as \texttt{PREVAP} and \texttt{PREVCHD},  excessive smoothing regularization can distort their true impact on survival outcomes. The shape functions for these rare categorical variables are vulnerable to being inappropriately regressed toward the population mean, causing established cardiovascular risk factors to paradoxically appear protective in the visualized contribution plots. This phenomenon occurs because the limited number of positive cases provides insufficient signal to overcome the model's smoothing penalties, resulting in misleading shape functions that fail to capture the true elevated risk associated with these clinical conditions.
    \item \textbf{Selection.}  The original study excluded most people with severe existing heart disease.  The retained group is therefore healthier or already under treatment, a “survival‑selection’’ bias that lowers their short‑term risk estimates~\cite{DAgostino1989FHS}.
    \item \textbf{Competing‑risk censoring.}  Deaths due to heart problems are counted under \textit{Risk 1}.  Removing those events from the \textit{Risk 1} pool leaves a group that is, by definition, less likely to die from non‑heart causes.  This negative effect can then bleed back into the \textit{Risk 1} estimate because the true positive effect must be learned from very few events.
\end{enumerate}

For \textit{Risk 2}, the shape functions for \texttt{SYSBP} and \texttt{GLUCOSE} follow J-shaped profiles, with minimal contribution at lower values and a marked increase beyond the upper quantiles. \texttt{educ} decreases nearly linearly. The binary variable \texttt{PREVCHD} is associated with a negative contribution, while \texttt{DIABETES} presents a discrete step-wise increase. The \texttt{BMI} function shows a shallow U-shape. \texttt{DIABP} increases monotonically throughout its observed range.

 Several observed patterns in the shape functions are consistent with known risk factors for cardiovascular and all-cause mortality. Established studies have linked elevated systolic blood pressure and high glucose levels with heightened cardiovascular risk~\citep{kannel1979diabetes, DAgostino1989FHS}. Cigarettes per day displayed an exponential relationship with diminishing marginal effects at higher consumption levels, reflecting saturation of smoking-related harm pathways. The inverse trend for educational attainment aligns with literature on socioeconomic disparities in cardiovascular outcomes~\citep{rosengren2019socioeconomic}. The U-shaped relationship observed for BMI (Risk 2) has been previously noted in older adults and is often described as the “obesity paradox”~\citep{amundson2010obesity} as well as sparse data in the extreme ranges.  Additionally, previous cardiovascular conditions (angina pectoris and coronary heart disease) showed positive linear associations with mortality risk. 

\begin{figure}[htbp]
    \centering
    \begin{subfigure}[ht]{0.48\textwidth}
        \includegraphics[width=\linewidth, height=8cm, keepaspectratio]{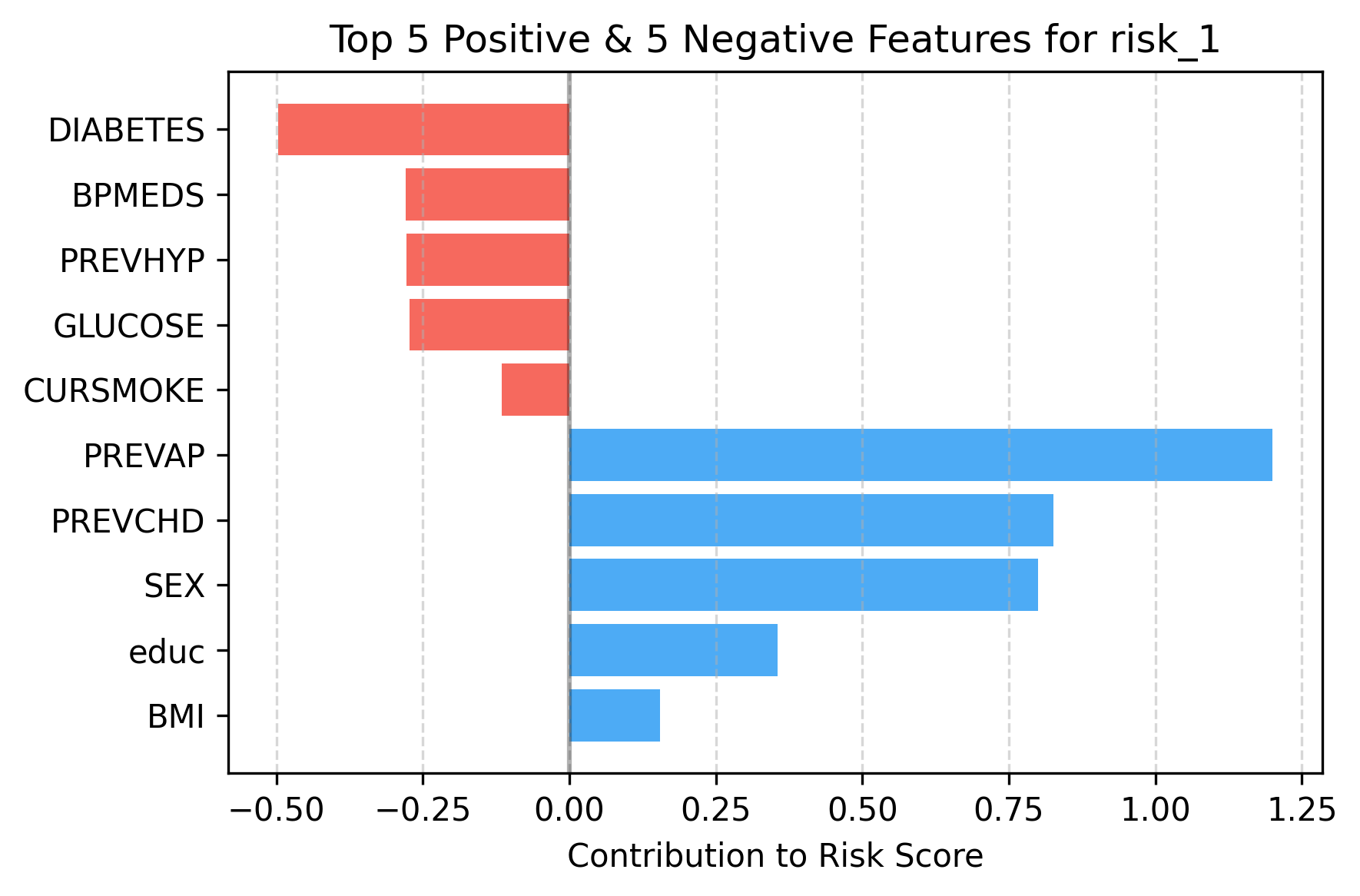}
    \end{subfigure}
    \hfill
    \begin{subfigure}[ht]{0.48\textwidth}
        \includegraphics[width=\linewidth, height=8cm, keepaspectratio]{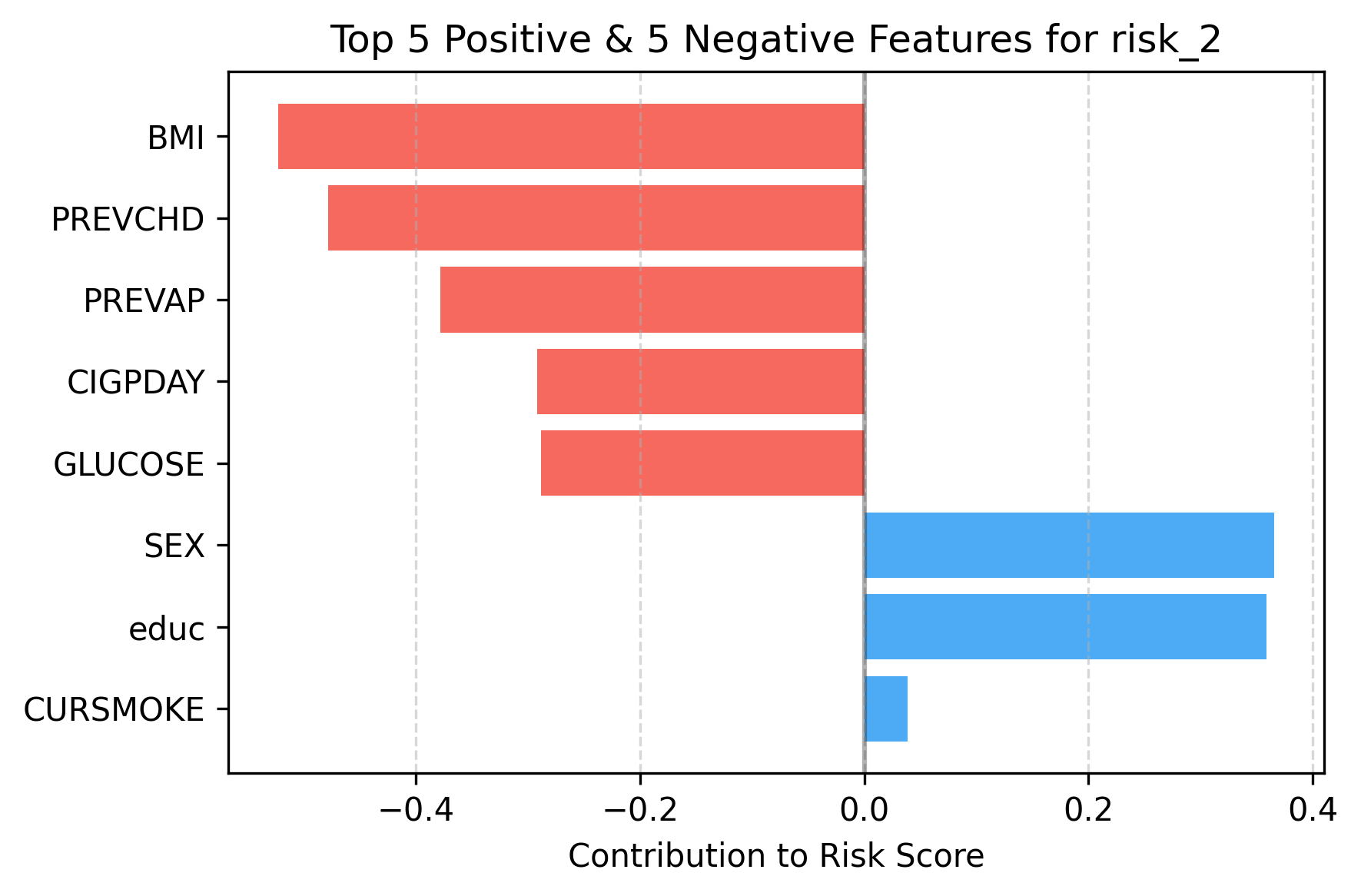}
    \end{subfigure}
    \caption{Feature Importances computed using CRISP-NAM for the Framingham Dataset.}
    \label{fig:feature_importance_framingham}
\end{figure}

Figure~\ref{fig:feature_importance_framingham} shows how features contribute positively or negatively to the prediction. While feature importance plots are generally not as informative as shape plots, which can reveal more detailed relationship between covariates, we provide it here for completeness.

\subsubsection{Primary Biliary Cholangitis (PBC) Dataset}

\begin{figure}[htbp]
    \centering
    \begin{subfigure}[b]{\textwidth}
        \includegraphics[width=\textwidth, height=8cm, keepaspectratio]{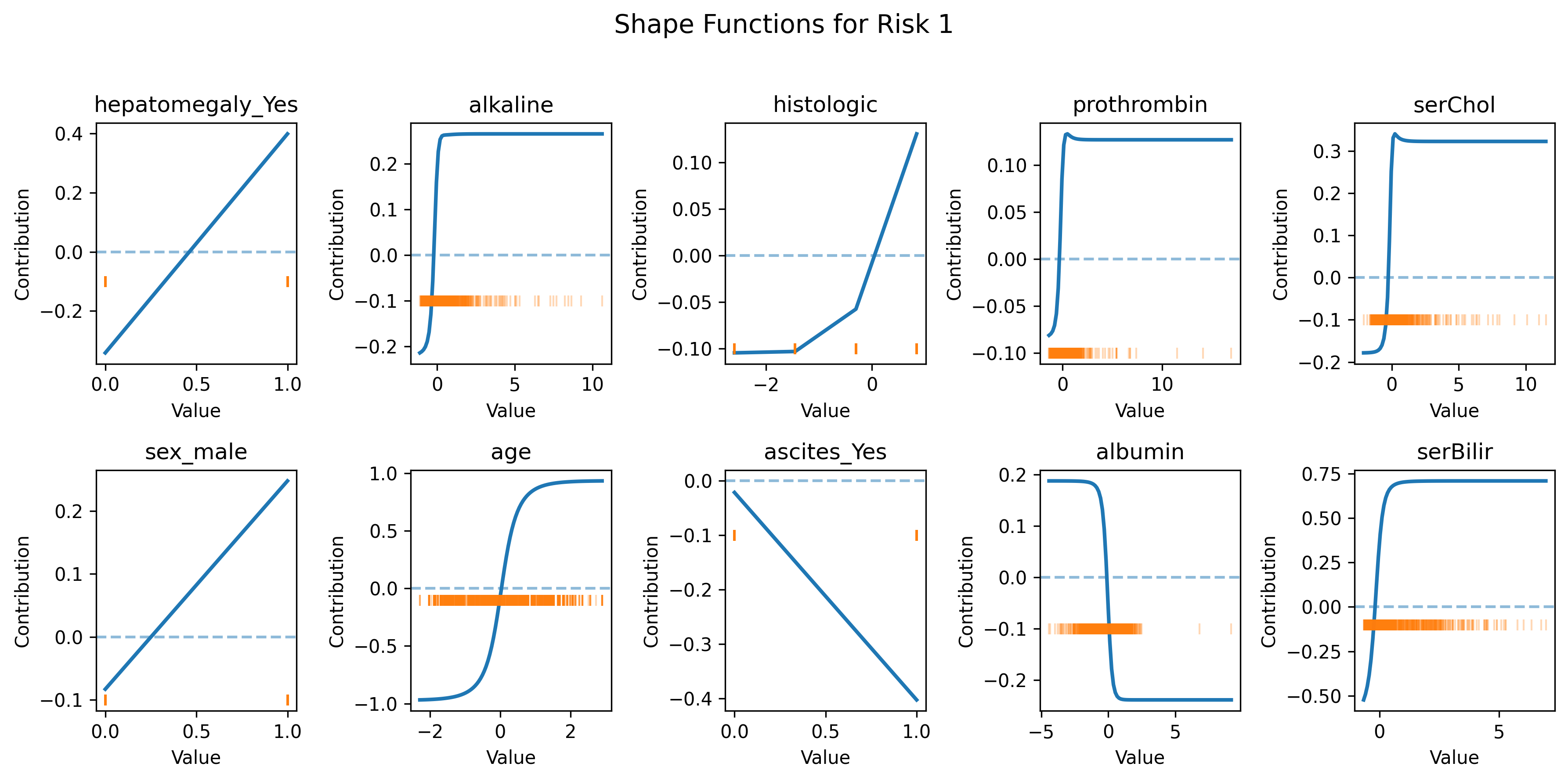}
    \end{subfigure}
    \vspace{1em}
    \begin{subfigure}[b]{\textwidth}
        \includegraphics[width=\textwidth, height=8cm, keepaspectratio]{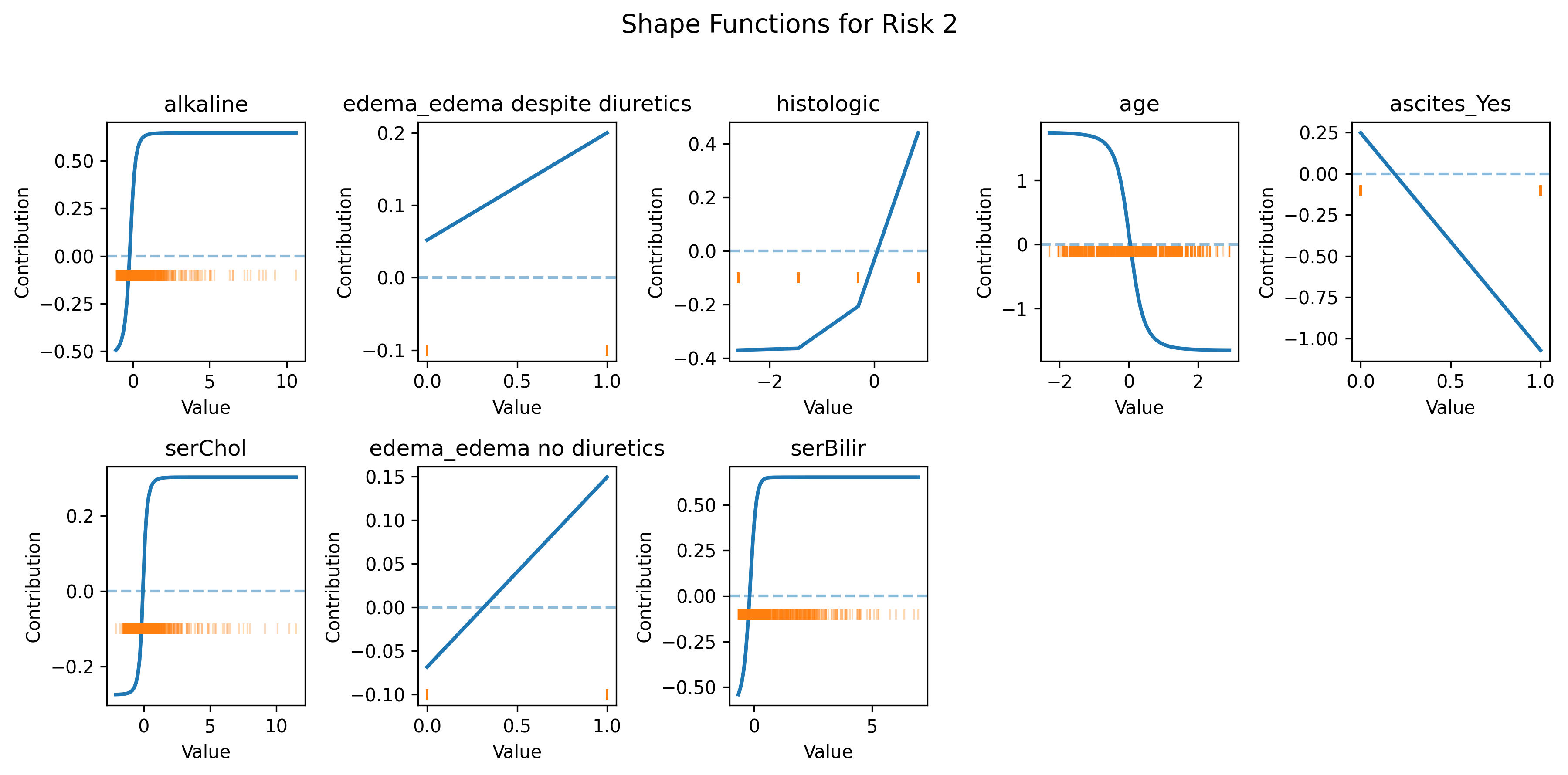}
    \end{subfigure}
    \caption{Shape functions computed with CRISP-NAM model for the top-10 important features: PBC dataset}
    \label{Fig:PBC_shape}
\end{figure}

\begin{figure}[htbp]

    \centering
    \begin{subfigure}[b]{0.48\textwidth}
        \includegraphics[width=\linewidth, height=8cm, keepaspectratio]{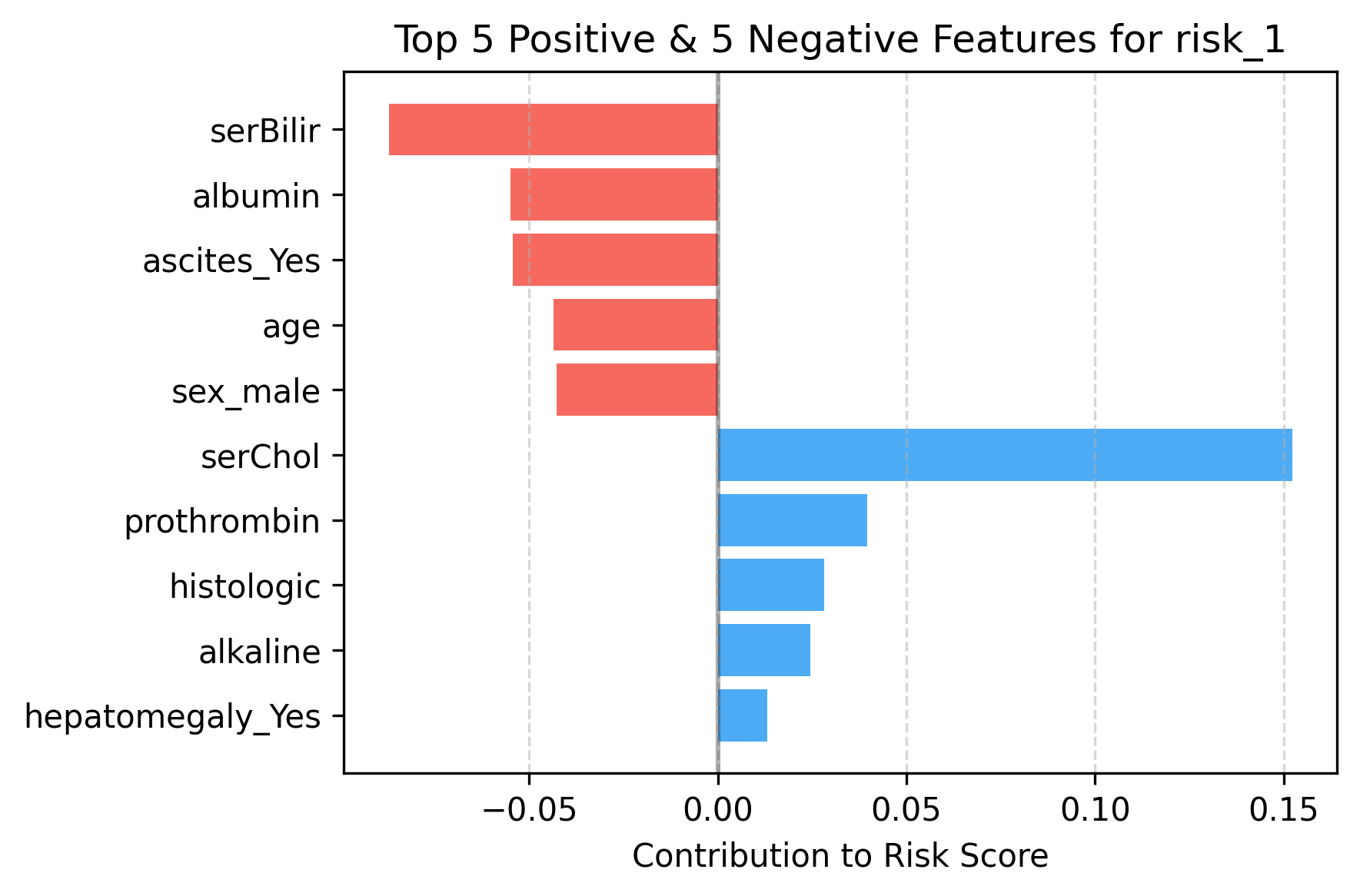}
    \end{subfigure}
    \hfill
    \begin{subfigure}[b]{0.48\textwidth}
        \includegraphics[width=\linewidth, height=8cm, keepaspectratio]{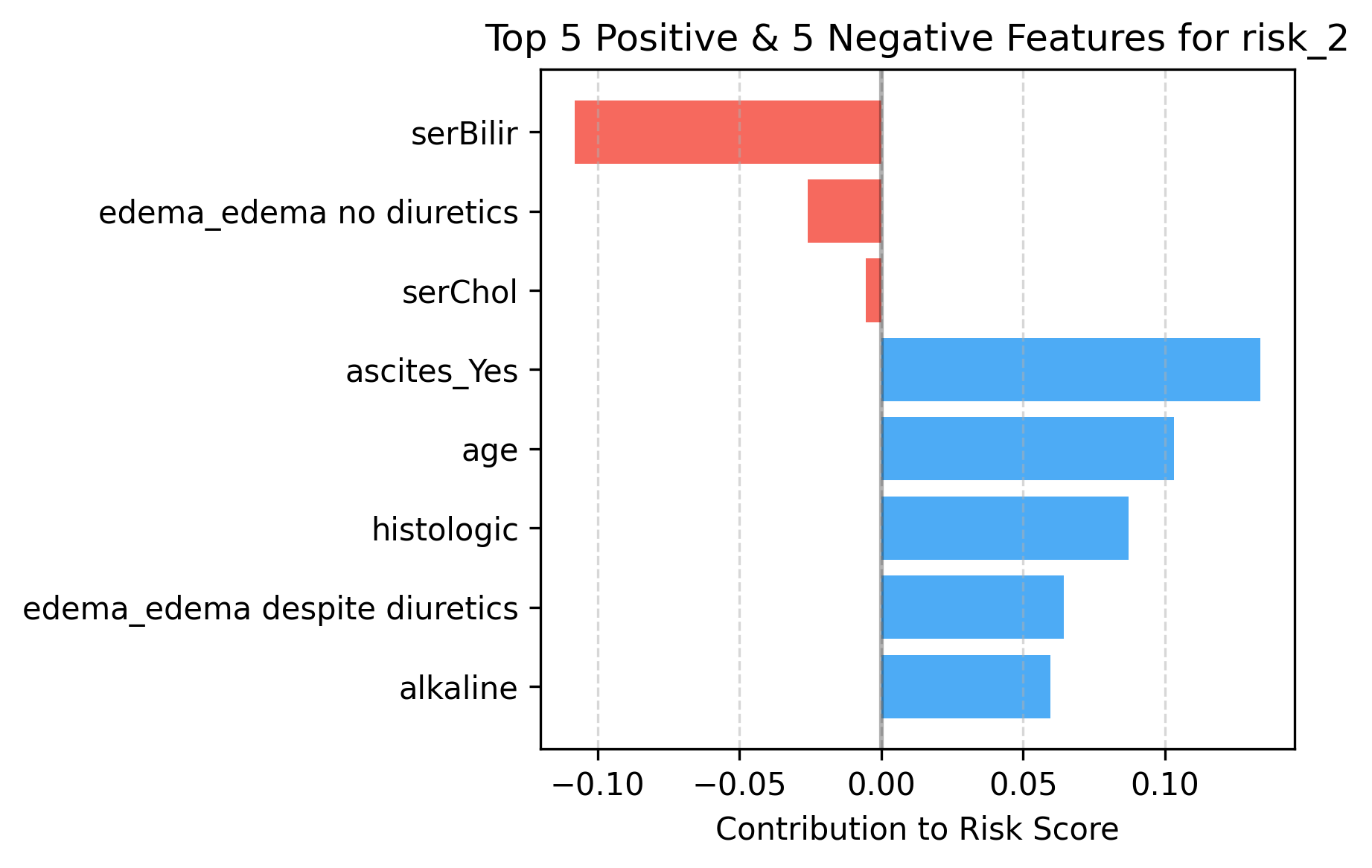}
    \end{subfigure}
    \caption{Feature Importances computed using CRISP-NAM for the PBC Dataset.}
    \label{fig:feature_importance_pbc}
\end{figure}

Figure~\ref{Fig:PBC_shape} presents the feature importance and shape functions from the CRISP-NAM model trained on the Primary Biliary Cholangitis (PBC) dataset. The plots show distinct patterns for \textit{Risk 1} (death on the waiting list) and \textit{Risk 2} (transplantation). Additionally, Figure~\ref{fig:feature_importance_pbc} shows top 5 positive and top 5 negative important features that contributed to the model's prediction.

For \textit{Risk 1}, \texttt{age} exhibits a Sigmoid-shaped curve. \texttt{alkaline} displays sharp increases followed by plateaus at higher values.Biomarkers \texttt{serBilir}, \texttt{serChol} and \texttt{prothrombin} all show similar rises steeply at low values and plateus thereafter. Binary indicators such as \texttt{hepatomegaly\_Yes} shows positive contributions. \texttt{albumin} demonstrates an inverted Sigmoid-shaped curve: initially increasing, crossing zero near the median, and declining at higher values. 

For \textit{Risk 2}, the shape for \texttt{age} exhibits inverse sigmoidal shape with a decreasing trend as age increases signifying that transplantation risk decreases as patients progress with age.  The shape for \texttt{serBilir}, \texttt{alkaline} and \texttt{serChol} rises steeply, before plateauing. These three elevated biomarkers indicate disease severity in PBC, which simultaneously increases both death risk and transplant priority.  Examining the contribution scales for both risks for \texttt{ascites\_Yes} shows a moderate negative contribution to Risk 1 ($\sim{-0.4}$) but a much stronger negative contribution to Risk 2 ($\sim{-1.0}$). From a model validation perspective, this suggests the model has learned that ascites presence is associated with reduced likelihood of both outcomes, but particularly transplantation. The asymmetric magnitudes indicate the model distinguishes between the two competing risks rather than simply treating ascites as a general severity marker.

The rug plots accompanying each shape function reflect the distribution of the feature values and indicate regions with limited data support. These empirical patterns align with several well-established clinical insights in the context of Primary Biliary Cholangitis (PBC). For instance, older age, elevated liver enzymes such as alkaline phosphatase, and increased serum bilirubin are recognized markers of disease severity and poorer prognosis~\citep{Dickson1989, MurilloPerez2023}. The decreasing transplant hazard with age may reflect clinical prioritization criteria that favour younger candidates for organ allocation. The non-linear shape for albumin aligns with its known role as a proxy for liver synthetic function, where low levels indicate hepatic decompensation. Histologic stage (\texttt{histologic}) progression, from fibrosis to cirrhosis, is a standard determinant in transplant eligibility, consistent with the monotonic rise observed in its shape function. Additionally, ascites and hepatomegaly are classical signs of advanced liver disease, often associated with higher mortality and reduced transplant suitability. Edema in PBC patients indicates advanced liver disease. In the competing risks framework, patients with edema have higher disease severity and thus receive higher priority for transplantation due to their urgent medical need. The positive contribution to transplantation risk reflects the clinical reality that transplant allocation systems prioritize sicker patients and those with edema are more likely to receive transplants because it serves as a marker of advanced disease requiring urgent intervention.

\subsubsection{SUPPORT2 Dataset}

\begin{figure}[htbp]
    \centering
    \begin{subfigure}[b]{\textwidth}
        \includegraphics[width=\textwidth, height=8cm, keepaspectratio]{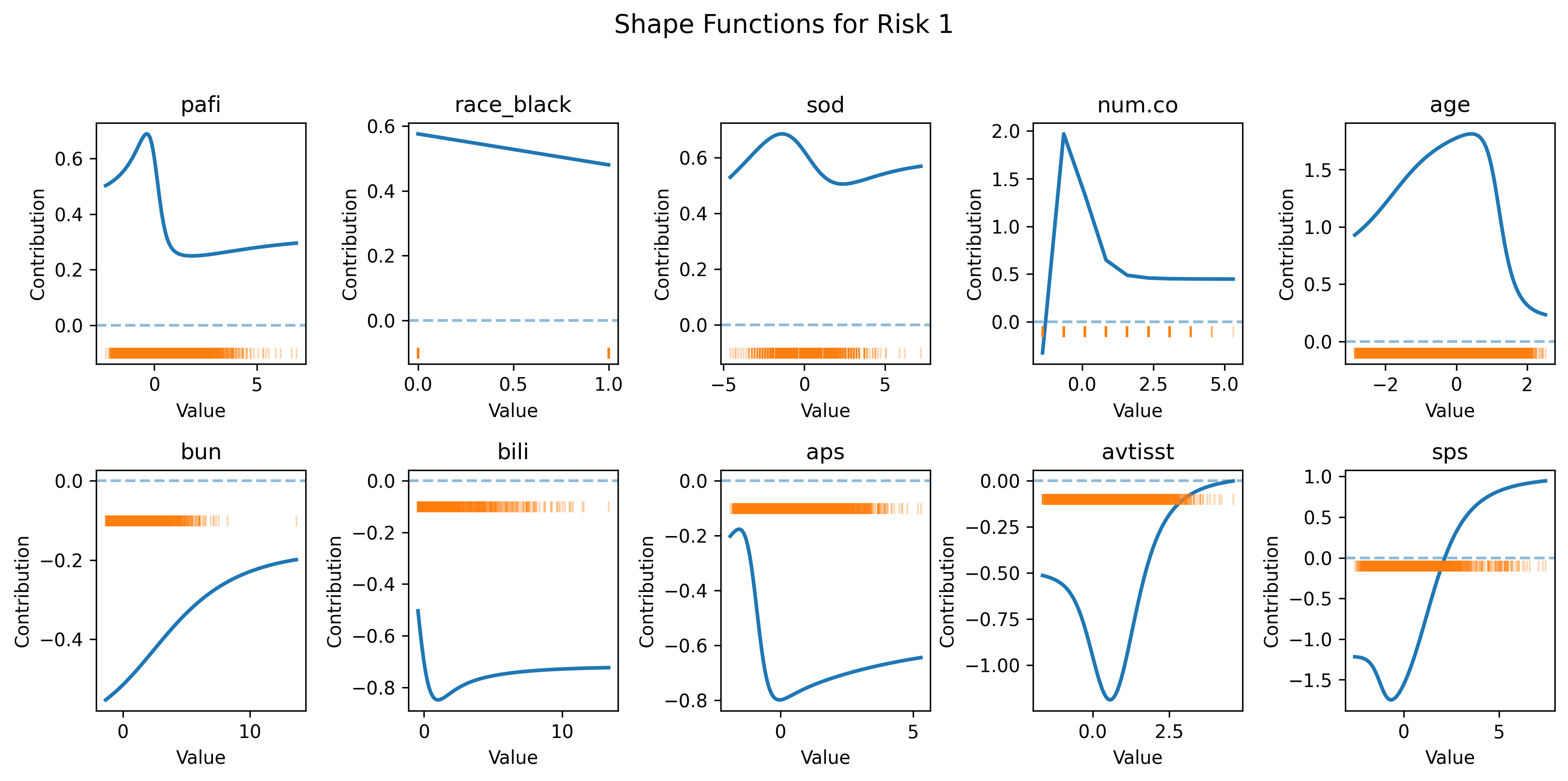}
    \end{subfigure}
    \vspace{1em}
    \begin{subfigure}[b]{\textwidth}
        \includegraphics[width=\textwidth, height=8cm, keepaspectratio]{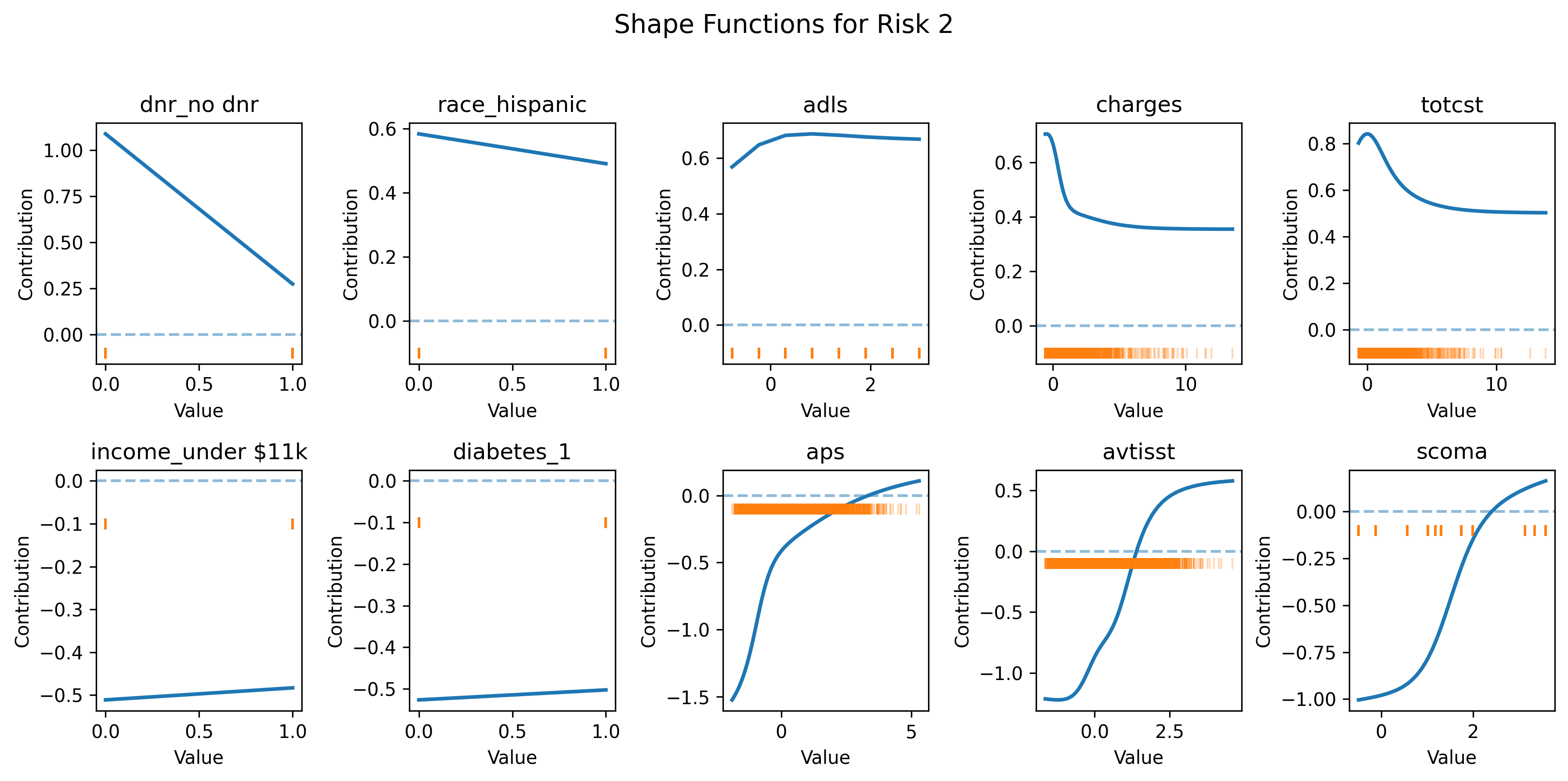}
    \end{subfigure}
    \caption{Shape functions computed with CRISP-NAM model for the top-10 important features: SUPPORT2 dataset}
    \label{Fig:SUPPORT2_shape}
\end{figure}

Figure~\ref{Fig:SUPPORT2_shape} presents feature importance and shape functions from the CRISP-NAM model trained on the SUPPORT2 dataset, which distinguishes cancer-specific mortality (\textit{Risk 1}) from death due to other causes (\textit{Risk 2}). 

We highlight several observations from the shape plots for both risks. 
For \textit{Risk 1}, the shape function for \texttt{age} shows a lower risk of death from cancer for younger patients, then steadily increasing risk of cancer-related death up to approximately 65 years and declining risk for very old patients. The binary indicator \texttt{race\_black} exhibits a slight negative contribution to the log-hazard for Risk 1 showing that black patients appear to have lower mortality due to cancer. The apparent protective effect of Black race contradicts established epidemiological evidence~\citep{saka2025cancer} demonstrating higher cancer mortality rates in Black population. This could indicate  insufficient sample representation or selection biases inherent to the SUPPORT2 dataset.  Similarly, the inverse relationship between number of comorbidities and cancer death risk suggests competing mortality mechanisms, where patients with multiple comorbidities may succumb to other medical conditions before cancer progression becomes the primary threat. The shape for \texttt{avtisst} is  lower durations of mechanical ventilation but shows a steep increase beyond certain number of days. \texttt{pafi} (oxygen ratio) exhibits a spike at very low values ($\sim{0.7}$), then drops to steady negative contribution ($\sim{0.3}$). Other biomarkers such as sodium \texttt{sod} shows an inverted U-shape peaking near normal levels, suggesting both low sodium levels as well as very high sodium levels increases risk.

For \textit{Risk 2}, the severity scores \texttt{aps} show sharp increases followed by plateaus. The \texttt{avtisst} variable again displays a marked rise in hazard for durations exceeding 3 days. Cost-related variables \texttt{charges} and \texttt{totcst} show decreasing trends. The binary indicator \texttt{dnr\_no\_dnr} shows that patients without DNR (Do Not Resuscitate)  orders (\texttt{dnr\_no\_dnr = 1}) demonstrated substantially lower contributions to non-cancer death risk compared to those with DNR orders present. This pattern aligns with clinical expectations, as DNR orders typically indicate patients with poor overall prognosis, advanced chronic diseases, or end-stage conditions who are at elevated risk for cardiovascular, respiratory, or multi-organ failure.  Rising  \texttt{aps} scores are consistent with the role of physiological instability and organ dysfunction in predicting mortality~\citep{Knaus1995support}. The inverted-U for \texttt{age} in cancer mortality likely reflects competing risks from other causes in older individuals~\citep{hayes2020competing}. The association between prolonged ventilation (\texttt{avtisst} $>3$ days) and increased hazard aligns with the known severity of illness in patients requiring extended respiratory support. Cost variables likely act as proxies for length of stay or illness trajectory rather than direct predictors.

\begin{figure}[ht!]
\label{fig:feature_importance_support}
    \centering
    \begin{subfigure}[ht]{0.48\textwidth}
        \includegraphics[width=\linewidth, height=8cm, keepaspectratio]{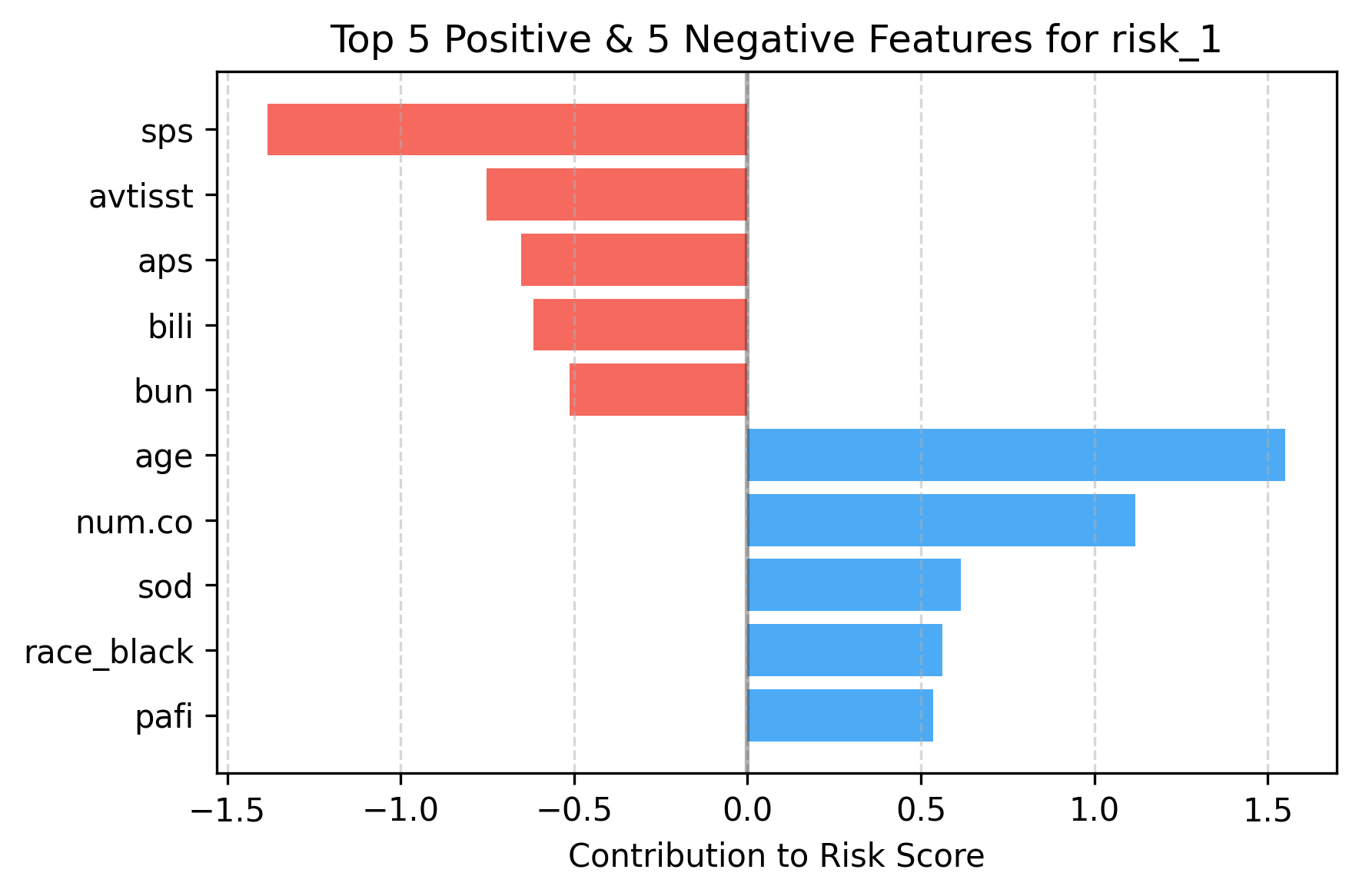}
    \end{subfigure}
    \hfill
    \begin{subfigure}[ht]{0.48\textwidth}
        \includegraphics[width=\linewidth, height=8cm, keepaspectratio]{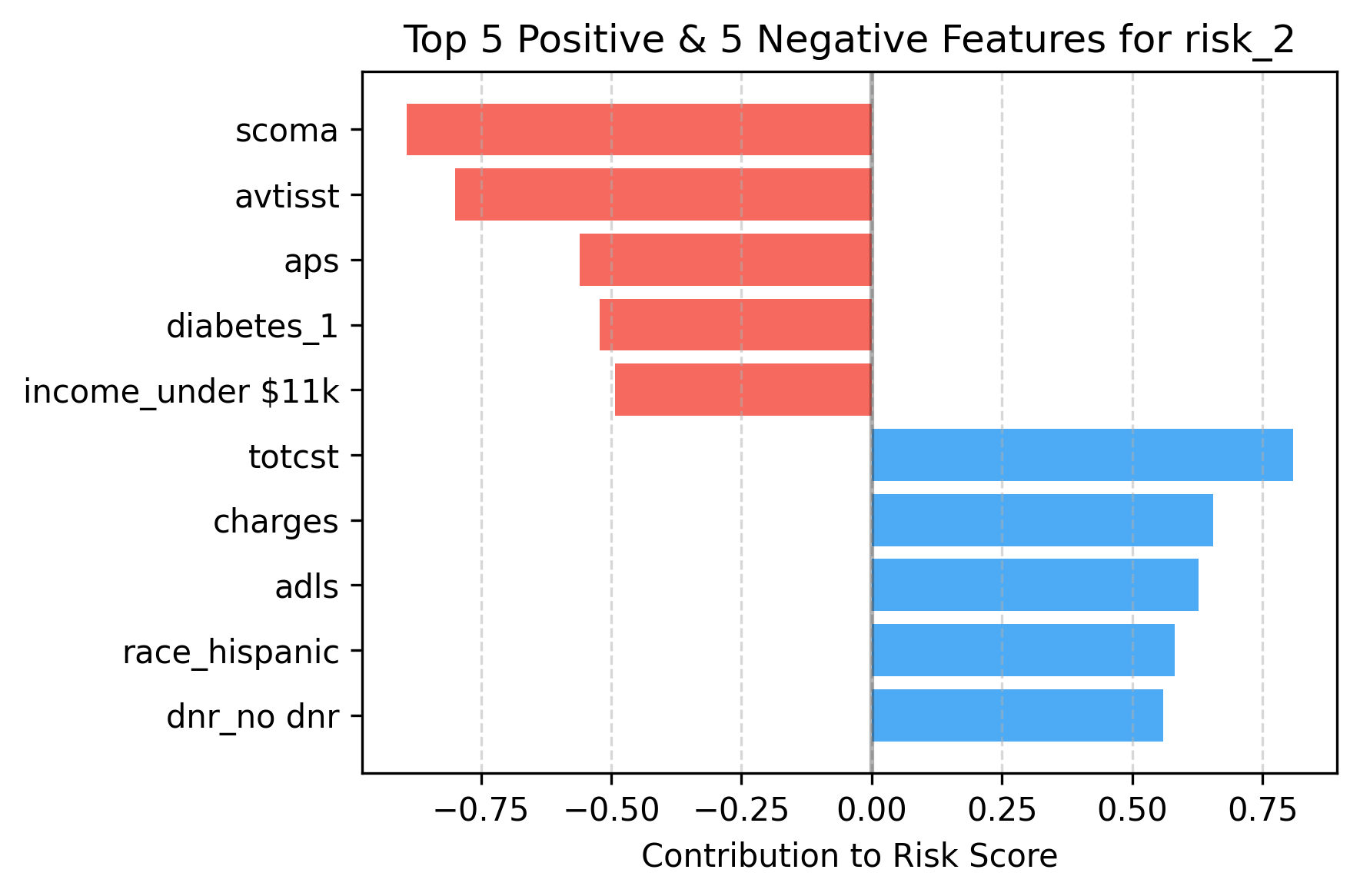}
    \end{subfigure}
    \caption{Feature Importances computed using CRISP-NAM for the SUPPORT2 Dataset.}
\end{figure}

\section{Conclusion}

We introduce CRISP-NAM, a deep survival model that simultaneously addresses competing risks in survival analysis while remaining inherently interpretable.  Our model demonstrates competitive discriminative performance on real-world clinical data and uniquely reveals covariate effects through intuitive shape function plots. CRISP-NAM is particularly valuable in high-stakes healthcare ML applications requiring mechanistic understanding such as investigating associational relationships, assessing treatment efficacy, or designing targeted interventions, especially when competing events represent distinct clinical processes~\citep{austin2016introduction}.  

We acknowledge that CRISP-NAM inherits the proportional hazards assumption from the Cox framework, requiring that covariate effects on each cause-specific hazard remain constant over time. This assumption may be violated when covariate effects vary temporally, such as biomarkers having different predictive power for early versus late events. There are two  possible avenues for future work: (i) exploration of temporal FeatureNets capable of learning how feature contributions evolve over time. The caveat that is that this approach would likely increase computational complexity during training and potentially demand larger datasets. Additionally, we will investigate using a modified loss function that takes calibration (Brier Score) into account.

\bibliography{references}
\bibliographystyle{plainnat}

\end{document}

%% file: arch-diag.tex
\usetikzlibrary{positioning, calc, shapes.multipart}

\begin{tikzpicture}[
  scale=0.75, transform shape,
  feature/.style={rectangle, draw=blue!70, fill=blue!10, rounded corners=3pt, minimum width=2.5cm, minimum height=1cm, font=\large},
  featurenet/.style={rectangle, draw=green!70, fill=green!10, rounded corners=5pt, minimum width=3.5cm, minimum height=1cm, font=\large},
  representation/.style={rectangle, draw=orange!70, fill=orange!10, rounded corners=3pt, minimum width=1.7cm, minimum height=1cm, font=\large},
  riskproj/.style={rectangle, draw=black!70, fill=white, rounded corners=3pt, minimum width=3.5cm, minimum height=0.6cm, font=\large},
  sumcircle/.style={circle, draw=black, fill=gray!10, minimum size=0.9cm, font=\large},
  riskagg/.style={rectangle, draw=gray!80, fill=gray!10, rounded corners=5pt, minimum width=3.2cm, minimum height=1cm, font=\large, font=\bfseries},
  riskscore/.style={rectangle, draw=blue!70, fill=blue!10, rounded corners=3pt, minimum width=2.3cm, minimum height=1cm, font=\large},
  riskscore2/.style={rectangle, draw=red!70, fill=red!10, rounded corners=3pt, minimum width=2.3cm, minimum height=1cm, font=\large},
  hazardsection/.style={rectangle, draw=yellow!70, fill=yellow!10, rounded corners=5pt, minimum width=22cm, minimum height=4.5cm, font=\large},
  baseline1/.style={rectangle, draw=blue!70, fill=blue!10, rounded corners=3pt, minimum width=3cm, minimum height=1cm, font=\large},
  baseline2/.style={rectangle, draw=red!70, fill=red!10, rounded corners=3pt, minimum width=3cm, minimum height=1cm, font=\large},
  hazard1/.style={rectangle, draw=blue!70, fill=blue!10, rounded corners=3pt, minimum width=3cm, minimum height=1cm, font=\large},
  hazard2/.style={rectangle, draw=red!70, fill=red!10, rounded corners=3pt, minimum width=3cm, minimum height=1cm, font=\large},
  cif1/.style={rectangle, draw=blue!70, fill=blue!10, rounded corners=3pt, minimum width=3cm, minimum height=1cm, font=\large},
  cif2/.style={rectangle, draw=red!70, fill=red!10, rounded corners=3pt, minimum width=3cm, minimum height=1cm, font=\large},
  formula/.style={rectangle, draw=gray!50, fill=white, rounded corners=3pt, font=\large},
  >=stealth,
]

\node[font=\Large\bfseries] at (0, 0) (title) {CRISP-NAM: Model Architecture};

\node[feature, below=1.5cm of title] (x1) {Feature $x_1$};
\node[feature, below=2.0cm of x1] (x2) {Feature $x_2$};
\node[below=0.5cm of x2] (dots1) {$\vdots$};
\node[feature, below=1.1cm of dots1] (xp) {Feature $x_p$};

\node[featurenet, right=1.5cm of x1] (f1) {FeatureNet $f_1(x_1)$};
\node[featurenet, right=1.5cm of x2] (f2) {FeatureNet $f_2(x_2)$};
\node[right=1.5cm of dots1] (dots2) {$\vdots$};
\node[featurenet, right=1.5cm of xp] (fp) {FeatureNet $f_p(x_p)$};

\node[representation, right=1.5cm of f1] (h1) {$\mathbf{h}_1$};
\node[representation, right=1.5cm of f2] (h2) {$\mathbf{h}_2$};
\node[right=1.5cm of dots2] (dots3) {$\vdots$};
\node[representation, right=1.5cm of fp] (hp) {$\mathbf{h}_p$};

\node[riskproj, above right=0.1cm and 1.0cm of h1] (g11) {$g_{1,1}(\mathbf{h}_1)$};
\node[riskproj, above right=0.1cm and 1.0cm of h2] (g21) {$g_{2,1}(\mathbf{h}_2)$};
\node[riskproj, above right=0.1cm and 1.0cm of hp] (gp1) {$g_{p,1}(\mathbf{h}_p)$};
\node[font=\bfseries\Large, above=0.1cm of g11.north] {Projections};

\node[riskproj, below right=0.1cm and 1.0cm of h1] (g12) {$g_{1,2}(\mathbf{h}_1)$};
\node[riskproj, below right=0.1cm and 1.0cm of h2] (g22) {$g_{2,2}(\mathbf{h}_2)$};
\node[riskproj, below right=0.1cm and 1.0cm of hp] (gp2) {$g_{p,2}(\mathbf{h}_p)$};

\node[sumcircle, right=2.8cm of g12] (sum1) {+};
\node[sumcircle, right=2.8cm of g22] (sum2) {+};

\node[riskagg, right=1.2cm of sum1] (agg1) {Additive Aggregator};
\node[riskagg, right=1.2cm of sum2] (agg2) {Additive Aggregator};


\node[riskscore, right=1.2cm of agg1] (eta1) {$\eta_1(\mathbf{x})$};
\node[riskscore2, right=1.2cm of agg2] (eta2) {$\eta_2(\mathbf{x})$};
\node[font=\bfseries\Large, above=0.1cm of eta1.north] {Cause-specific log-hazard};

\node[formula, below=3.8cm of eta2,xshift=-2cm] (formula) {$\lambda_k(t|\mathbf{x}) = \lambda_{0k}(t) \cdot \exp(\eta_k(\mathbf{x}))$};
\node[hazardsection, below=2.5cm of formula.south] (hazardbox) {};
\node[font=\bfseries\Large, below=0.1cm of hazardbox.south] {Baseline Hazard Estimation \& CIF Calculation};

\node[baseline1, left=7cm of hazardbox.center, yshift=1cm] (lambda01) {$\lambda_{01}(t)$};
\node[baseline2, left=7cm of hazardbox.center, yshift=-1cm] (lambda02) {$\lambda_{02}(t)$};

\node[hazard1, left=1.5cm of hazardbox.center, yshift=1cm] (lambda1) {$\lambda_1(t|\mathbf{x})$};
\node[hazard2, left=1.5cm of hazardbox.center, yshift=-1cm] (lambda2) {$\lambda_2(t|\mathbf{x})$};

\node[formula, right=2cm of lambda1] (cif1formula) {$F_1(t|\mathbf{x}) = \int_0^t S(u|\mathbf{x})\lambda_1(u|\mathbf{x})du$};
\node[formula, right=2cm of lambda2] (cif2formula) {$F_2(t|\mathbf{x}) = \int_0^t S(u|\mathbf{x})\lambda_2(u|\mathbf{x})du$};

\node[cif1, right=7cm of hazardbox.center, yshift=1cm] (F1) {$F_1(t|\mathbf{x})$};
\node[cif2, right=7cm of hazardbox.center, yshift=-1cm] (F2) {$F_2(t|\mathbf{x})$};

\foreach \i/\f/\h in {x1/f1/h1, x2/f2/h2, xp/fp/hp} {
  \draw[->] (\i) -- (\f);
  \draw[->] (\f) -- (\h);
}

\foreach \h/\gA/\gB in {h1/g11/g12, h2/g21/g22, hp/gp1/gp2} {
  \draw[->, blue!70, thick] (\h) -- (\gA.west);
  \draw[->, red!70, thick] (\h) -- (\gB.west);
}

\foreach \g in {g11, g21, gp1} {
  \draw[->, blue!70, thick] (\g.east) -- (sum1);
}
\draw[->, blue!70, thick] (sum1) -- (agg1);
\draw[->, blue!70, thick] (agg1) -- (eta1);

\foreach \g in {g12, g22, gp2} {
  \draw[->, red!70, thick] (\g.east) -- (sum2);
}
\draw[->, red!70, thick] (sum2) -- (agg2);
\draw[->, red!70, thick] (agg2) -- (eta2);

\draw[->, dashed] (eta1.east) -- ++(+2, 0) |- (formula);
\draw[->, dashed] (eta2.east) -- ++(+2, 0) |- (formula);

\draw[->, blue!70, thick] (lambda01) -- (lambda1);
\draw[->, red!70, thick] (lambda02) -- (lambda2);

\draw[->, blue!70, thick] (lambda1) -- (cif1formula);
\draw[->, red!70, thick] (lambda2) -- (cif2formula);

\draw[->, blue!70, thick] (cif1formula) -- (F1);
\draw[->, red!70, thick] (cif2formula) -- (F2);
\draw[->, dashed] (formula.south) -- ($(formula.south)!0.5!(hazardbox.north)$) -- (hazardbox.north);
\end{tikzpicture}